\documentclass[conference]{IEEEtran}
\usepackage{times}

\usepackage[numbers]{natbib}
\usepackage{multicol}
\usepackage[bookmarks=true]{hyperref}
\usepackage{xcolor}
\usepackage{multirow}
\usepackage{pifont}
\usepackage{amssymb}
\usepackage{amsmath}
\usepackage{import}
\usepackage{graphicx}
\usepackage{booktabs}
\usepackage{makecell}
\usepackage{tabularx}
\usepackage{bm}
\usepackage[normalem]{ulem}
\useunder{\uline}{\ul}{}
\usepackage[font=small,labelfont=normalfont]{caption}

\newcommand{\ours}{$\Psi_{0}$}

\newcommand{\redx}{\textcolor{red}{\ding{55}}}
\newcommand{\greencheck}{\textcolor{green!60!black}{\ding{51}}}

\begin{document}

\title{\ours: An Open Foundation Model Towards \\ Universal Humanoid Loco-Manipulation}


\author{
\authorblockN{
Songlin Wei\textsuperscript{1*},
Hongyi Jing\textsuperscript{1*},
Boqian Li\textsuperscript{1*},
Zhenyu Zhao\textsuperscript{1*},
Jiageng Mao\textsuperscript{1},
Zhenhao Ni\textsuperscript{1},
Sicheng He\textsuperscript{1},\\
Jie Liu\textsuperscript{1},
Xiawei Liu\textsuperscript{1},
Kaidi Kang\textsuperscript{1},
Sheng Zang\textsuperscript{1},
Weiduo Yuan\textsuperscript{1},
Marco Pavone\textsuperscript{2},
Di Huang\textsuperscript{3},
Yue Wang\textsuperscript{1$\dag$}
}
\authorblockA{
\textsuperscript{1}USC Physical Superintelligence (PSI) Lab~
\textsuperscript{2}NVIDIA~
\textsuperscript{3}WorldEngine\\
* Equal Contribution
$\dag$ Corresponding Author \\
\url{https://psi-lab.ai/Psi0}
}
}


%

\twocolumn[{
\renewcommand\twocolumn[1][]{#1}
\begin{center}
    \maketitle
    \centering
    \includegraphics[width=\textwidth]{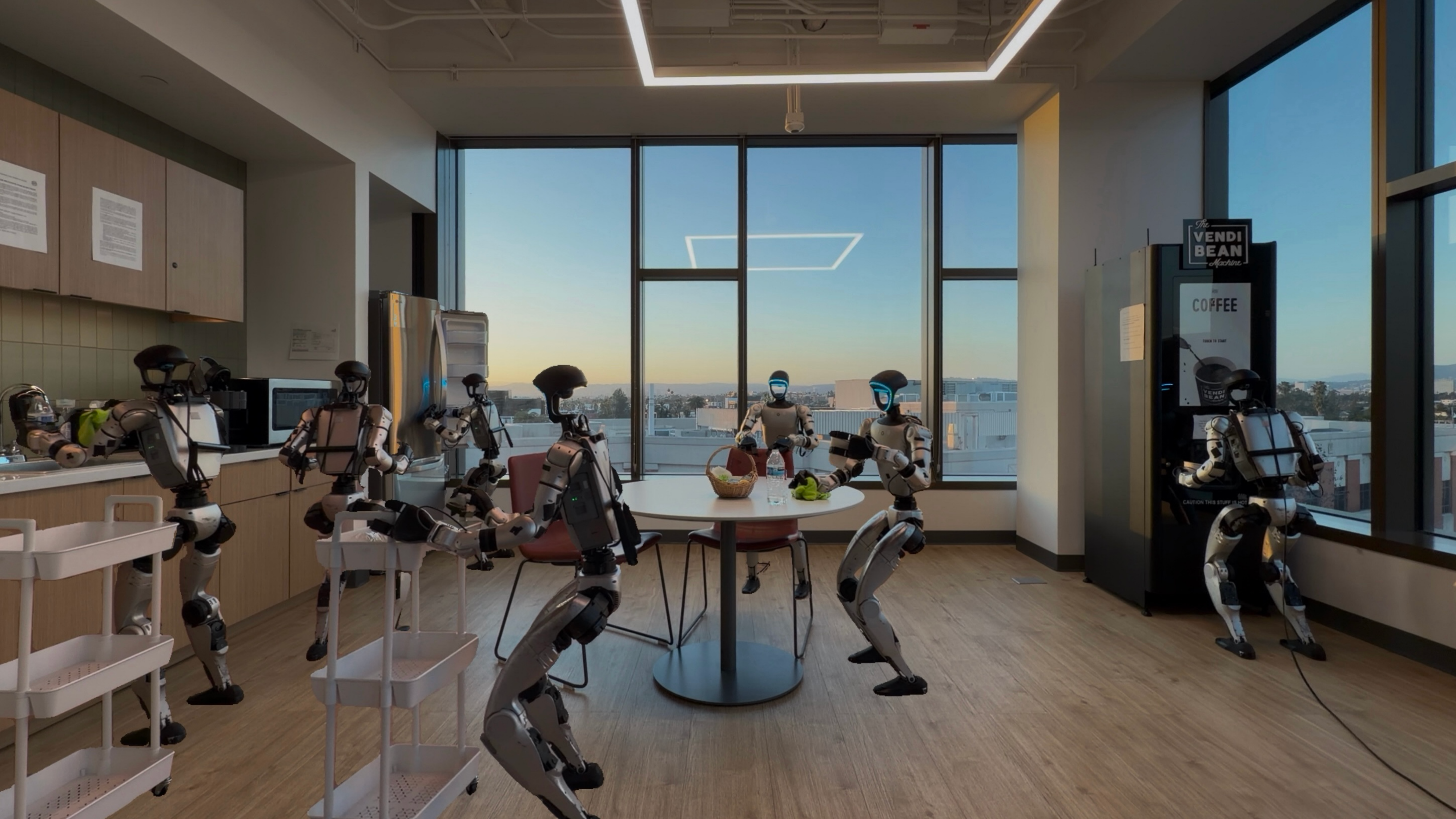}\\
    \vspace{0.2em}
    \small
    \captionsetup{type=figure}
    \captionof{figure}{\textbf{Humanoid Loco-Manipulation.} \ours~performs diverse loco-manipulation tasks in a pantry, including taking a cup from the coffee machine, pushing a cart, wiping the table, grasping a bottle and placing it in the sink, and pushing the fridge door.}
    \label{fig:teaser}
\end{center}
}]

\begin{abstract}
We introduce \ours\;(\textit{Psi-Zero}), an open foundation model to address challenging humanoid loco-manipulation tasks.
While existing approaches often attempt to address this fundamental problem by co-training on large and diverse human and humanoid data, we argue that this strategy is suboptimal due to the fundamental kinematic and motion disparities between humans and humanoid robots. Therefore, data efficiency and model performance remain unsatisfactory despite the considerable data volume. 
To address this challenge, \ours\;decouples the learning process to maximize the utility of heterogeneous data sources. 
Specifically, we propose a staged training paradigm with different learning objectives: First, we autoregressively pre-train a VLM backbone on large-scale egocentric human videos to acquire generalizable visual-action representations. Then, we post-train a flow-based action expert on high-quality humanoid robot data to learn precise robot joint control. 
 Our research further identifies a critical yet often overlooked data recipe: in contrast to approaches that scale with noisy Internet clips or heterogeneous cross-embodiment robot datasets, we demonstrate that pre-training on high-quality egocentric human manipulation data followed by post-training on domain-specific real-world humanoid trajectories yields superior performance.
Extensive real-world 
experiments demonstrate that \ours\ achieves the best performance using only about 800 hours of human video data and 30 hours of real-world robot data, outperforming baselines pre-trained on more than 10$\times$ as much data by over 40\% in overall success rate across multiple
tasks.
We will open-source the entire ecosystem to the community, including a data processing and training pipeline, a humanoid foundation model, and a real-time action inference engine. 
\end{abstract}

\IEEEpeerreviewmaketitle

\section{Introduction}
Humanoid robots, endowed with human-like morphology and dexterity, have achieved remarkable progress in whole-body motion control \cite{cheng2024expressive,liu2024visual, jiang2025wholebodyvla, ding2025humanoid}. 
However, their manipulation capabilities, which could eventually unlock enormous potential for society, have received less attention and faced greater challenges.
Recent advances in large language models (LLMs) have illuminated a promising path towards intelligence: by scaling both data and model capacity, general intelligence can emerge. Inspired by this paradigm, the robotics community has begun exploring scaling laws that are suitable for agents with physical bodies.
Recently, works such as RT 1-2 \cite{brohan2022rt, zitkovich2023rt}, OpenVLA \cite{kim2024openvla}, Gemini Robotics \cite{team2025gemini}, GR00T \cite{bjorck2025gr00t}, and
Physical Intelligence's $\pi_0, \pi_{0.5}$ \cite{black2024pi0, intelligence2025pi05} have advocated training large action models using massive amounts of real robot data. These approaches provide early evidence that the reasoning and planning abilities of large models can significantly improve generalization in robotic manipulation. 
However, these methods often rely on large-scale teleoperation data, which is prohibitively costly and challenging to acquire for humanoid loco-manipulation.

Fortunately, human egocentric videos provide a scalable alternative as they capture abundant natural motion patterns and rich behavior-level information without the expense of robot teleoperation.
However, directly transferring knowledge from human videos to humanoid control is non-trivial due to the substantial embodiment gap between humans and robots. Early efforts \cite{cai2025innon, yang2025egovla, bi2025hrdt} attempt to learn from human videos by adopting a unified human-centric state-action representation. 
Nevertheless, learning from such heterogeneous data remains challenging due to intrinsic discrepancies between humans and humanoids, including differences in action frequency, motion dynamics, and degrees of freedom. Although these approaches employ domain adaptation \cite{cai2025innon} or co-training strategies that mix human and robot data \cite{yang2025egovla}, a single monolithic policy that models two fundamentally different action distributions is inherently suboptimal. As a result, the learned policies still struggle to control humanoids to perform complex, long-horizon tasks. Therefore, this paper studies a fundamental question: \textit{how can we effectively distill motion priors and world knowledge from human egocentric videos to enable robust whole-body control for humanoid robots?}

To that end, we propose a novel multi-stage training paradigm with different learning goals for each stage: we first pre-train a VLM to predict next-step actions using the human-robot unified action space. 
The objective of this stage is to enable the model to learn task-level motion priors across diverse activities, while also learning visual representations aligned with downstream robotic tasks. 
We then train a separate flow-based action expert  using real humanoid robot data to predict action sequences directly in the joint space. This post-training stage includes both task-agnostic training on cross-task humanoid data and task-specific fine-tuning on in-domain teleoperated demonstrations.
We implement our action expert as a multi-modal diffusion transformer (MM-DiT) \cite{sd3}, which is more capable than a naive DiT.
Conditioned on the visual-language features from the VLM, the action expert efficiently and concurrently outputs joint-space action chunks. 
This stage enables the action expert to capture embodiment-specific dynamics. 
As a result, only a small amount of additional real-robot data is required for task-specific fine-tuning, after which the model can rapidly acquire long-horizon, dexterous loco-manipulation skills.

%

To enable effective training and deployment of our humanoid VLA, we make several key contributions. First, we optimize a manipulation-oriented teleoperation pipeline that improves lower-body stability during whole-body manipulation. 
Second, to ensure smooth execution in the real world at inference time, we introduce training-time real-time action chunking, which mitigates motion jitter caused by model inference latency. 
Finally, we deploy our model on a real humanoid robot and benchmark it against state-of-the-art methods on several complex, long-horizon tasks. 
Our experiments suggest that, using only 800 hours of human egocentric video and 30 hours of real-robot data, our model achieves significantly better performance than existing methods trained with more than 10 $\times$ as much data on long-horizon loco-manipulation tasks. These results reveal that \textbf{effective scaling requires scaling the right data in the right way.}
We will release the full training pipeline, pre-trained model weights, deployment code
to facilitate future research.
\section{Related Works}

\begin{figure*}[t]
  \centering
  \includegraphics[width=0.98\textwidth]{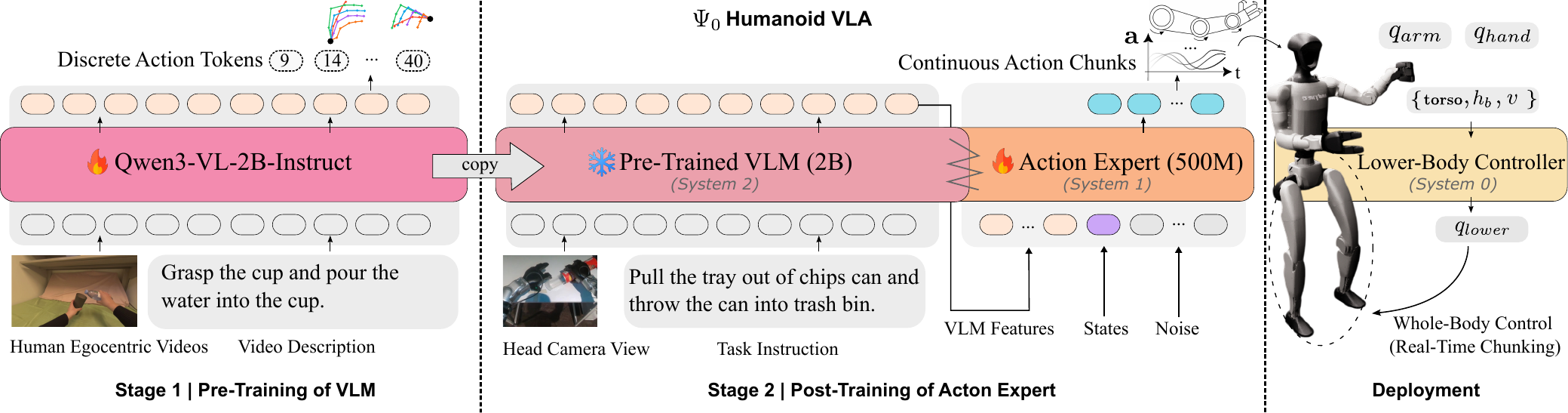}

\caption{\textbf{Model Training and Deployment}: First, we pre-train the VLM on the EgoDex \cite{egodex} dataset to autoregressively predict the next-action tokens in the task space. Then, we post-train the flow-based action expert using robotic data to predict action chunks in the joint space. Finally, we implement a real-time chunking mechanism that leverages the lower-body controller to achieve smooth whole-body control.}
\vspace{-1.5em}

  \label{fig:arch}
\end{figure*}
\subsection{Whole-Body Dexterous Manipulation}

Humanoid whole-body control has witnessed significant progress in recent works \cite{ze2025twist, cheng2024expressive, li2025clone, liao2025beyondmimic, fu2024humanplus, allshire2025visual, zhao2025resmimicgeneralmotiontracking, qi2025coordinated}. 
Humanoid robots are now able to mimic diverse human motions like running, dancing, and even flipping. 
Despite significant progress in locomotion, researchers have struggled to achieve comparable success in humanoid dexterous loco-manipulation.
LangWBC \cite{shao2025langwbc} and LeVERB \cite{xue2025leverb} introduce language-conditioned whole-body control policies, allowing humanoid robots to robustly execute high-level and language-specified behaviors. However, these methods primarily focus on locomotion and navigation, with limited emphasis on dexterous manipulation scenarios.
In parallel, AMO \cite{li2025amo} and TWIST2 \cite{ze2025twist2} enable humanoid whole-body control through VR-based teleoperation, providing an effective framework for collecting loco-manipulation data. However, they emphasize more on low-level control, rather than learning a precise policy for long-horizon dexterous loco-manipulation.


Dexterous manipulation \cite{grauman2024ego}, on the other hand, poses a long-standing challenge due to the high degree-of-freedom control and frequent self-occlusion between palms and fingers, which make vision-based dexterous manipulation extremely challenging.
Being-H0 \cite{luo2025being} proposes to learn from human video by curating a large amount of hand-object interaction videos and fine-tuning a pre-trained VLM using multiple task data like motion-infilling and translation. 
However, this method is limited to single-arm tabletop manipulation. 
To address the mentioned challenges, we propose to build a unified VLA model for humanoid whole-body dexterous manipulation.

\subsection{Humanoid VLAs}
Inspired by the remarkable success of foundation models, VLAs \cite{zitkovich2023rt, kim2024openvla, black2024pi0, zhang2024uni, geng2023sage, team2025gemini} have emerged as a promising direction toward bringing artificial intelligence into the physical world.
$\pi$ series \cite{black2024pi0, intelligence2025pi05} demonstrate exceptional generalization and robustness across challenging manipulation scenarios, including dual-arm and mobile manipulation.
GR00T \cite{bjorck2025gr00t} further open-sources the first foundation model for humanoid robots, trained on a large-scale mixture of real-world and synthetic data generated from videos. 
However, in contrast to them, we find that training on higher-quality data is more critical than simply scaling to large volumes of heterogeneous cross-embodiment data.
In this work, we explore a new paradigm for training humanoid VLAs that leverages large-scale human egocentric video data, complemented by a smaller amount of real robot interaction data.

\subsection{Learning From Egocentric Videos}
Data scarcity remains a fundamental constraint in training VLAs, as teleoperation data collection is less efficient and more expensive to scale.
In contrast, human video data contains rich prior knowledge of human–object interactions~\cite{uh1, kareer2024egomimicscalingimitationlearning, yu2025egomilearningactivevision}, providing a scalable alternative.
Recent approaches, such as EgoVLA \cite{yang2025egovla} and In-n-On \cite{cai2025innon}, co-train their models on human video and robot data to predict unified human wrist and hand actions, followed by inverse kinematics (IK) during inference to map these predictions to robot actions.
Similarly, H-RDT \cite{bi2025hrdt} trains a large diffusion transformer (DiT) to predict arm and hand actions in the end-effector space.
%
However, co-training the model end-to-end on a mixture of humanoid and non-humanoid robot data is suboptimal, as the model must simultaneously learn two fundamentally different action distributions. Instead, we identify a critical yet overlooked training recipe: after pre-training with next-action prediction to learn task semantics and visual representations, we post-train the action expert to directly model actions in the joint space, thereby avoiding the inefficiencies of co-training.
\section{The \ours \;Foundation Model}
In this section, we introduce \ours\;(\textit{Psi-Zero}), a VLA model for humanoid dexterous loco-manipulation. 
Given a natural language task instruction $\ell$ and the current observation $\mathbf{o}_t$, our model predicts the whole-body action chunk $\mathbf{a}_{t:t+H}$. 
The observation $\mathbf{o}_t$ contains the current head camera image $\mathbf{I}_t$ and the whole-body proprioceptive state $\mathbf{q}_t$, including upper joint state,  torso roll, pitch, yaw, and the base height.
The action $\mathbf{a} \in \mathbb{R}^{36}$ is defined as $\{\mathbf{q}_{hand},\mathbf{q}_{arm}, \mathbf{torso}_{rpy},h_b, v_x,v_y,v_{yaw}, p_{yaw}\}$, where $\mathbf{q}_{hand} \in \mathbb{R}^{14}$ and $\mathbf{q}_{arm}\in \mathbb{R}^{14}$ are the two hand and arm joints respectively, $\mathbf{torso}_{rpy} \in \mathbb{R}^{3}$ is the torso roll, pitch, yaw.
$h_b \in \mathbb{R}$ is the base height of the humanoid and $v_x, v_y \in \mathbb{R}$ are the horizontal linear velocities, and $v_{yaw} \in \mathbb{R}$ denotes angular velocity around the upward direction. $p_{yaw} \in \mathbb{R}$ is the target yaw rotation.
We employ an RL-based control policy~\cite{li2025amo} to control the lower body and torso joints throughout data collection and policy evaluation.


\subsection{Model Architecture}
\ours\;is a foundation model that adopts a \textit{triple-system} architecture, following prior work \cite{intelligence2025pi05, bjorck2025gr00t}. As shown in Fig.~\ref{fig:arch}, the high-level policy consists of two end-to-end–trained components: a vision–language backbone (\textit{system-2}) and a multi-modal diffusion transformer (MM-DiT) action expert (\textit{system-1}).
We use the state-of-the-art vision–language foundation model Qwen3-VL-2B-Instruct \cite{Qwen3-VL} as \textit{system-2}. The action expert is implemented as a flow-based MM-DiT inspired by Stable Diffusion~3 \cite{sd3}, containing approximately 500M parameters. Compared to a naive DiT-based action head, this design enables more efficient fusion of action and vision–language features.
Conditioned on hidden features from the VLM backbone, the action expert predicts future whole-body action chunks $\mathbf{a}_{t:t+H}$. The 8-DoF lower-body actions $\{\mathbf{torso}_{rpy}, h_b, v_x, v_y, v_{yaw}, p_{yaw}\}$ are passed to \textit{system-0}, a RL–based tracking policy. We adopt the off-the-shelf controller AMO \cite{li2025amo}, which maps these inputs to 15-DoF lower-body joint angles $\mathbf{q}_{lower} \in \mathbb{R}^{15}$, including 3 DoF waist and 12 DoF leg joint. 
Together with the 28-DoF upper-body joints $(\mathbf{q}_{arm}, \mathbf{q}_{hand})$, the system outputs 43-DoF actions for whole-body control.

\subsection{Training Recipe}
We present an efficient training recipe for learning humanoid loco-manipulation skills from both human videos and real robot data.
The overall training procedure consists of three stages: 
first, pre-training the VLM backbone on the large-scale high-quality and diverse human egocentric videos; 
second, post-training the flow-based action expert on cross-task real humanoid data; and third, fine-tuning the action expert using a small amount of in-domain task data, which enables rapid adaptation to new tasks.

\begin{figure}[t]
\includegraphics[width=0.95\linewidth]{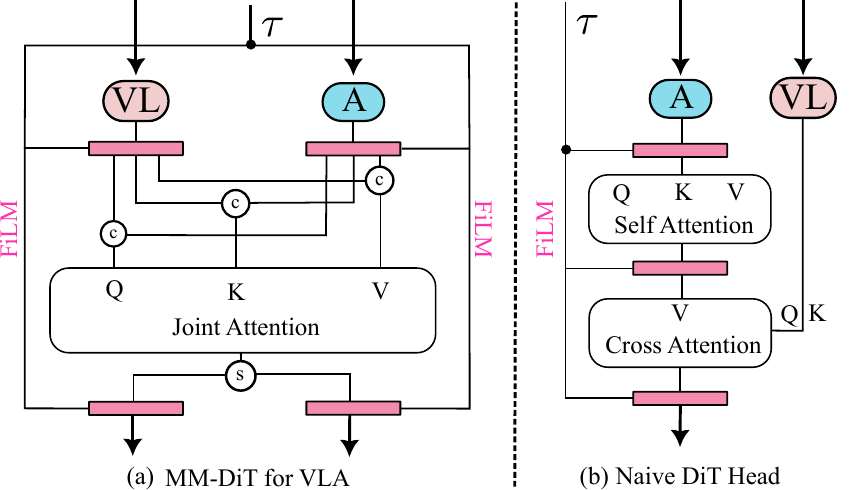}
\caption{\textbf{MM-DiT for VLA:} Comparison of MM-DiT architecture with naive DiT. $\tau$ is the flow timestep and \textbf{VL} and \textbf{A} denotes hidden states of the vision-language and action respectively.}
\vspace{-2em}
\label{fig:mm-dit}
\end{figure}

\subsubsection{Pre-Training on Egocentric Human Video}
Training a humanoid foundation model faces a significant data scarcity bottleneck. Human egocentric videos, which are much cheaper to scale than real-world robotics data, offer a promising alternative.
Therefore, we leverage EgoDex \cite{egodex}, which contains approximately 829 hours of human egocentric video capturing human hands performing diverse dexterous manipulation tasks.
To further mitigate the visual gap between human videos and robotic observations, we incorporate Humanoid Everyday \cite{he}, which contains 31 hours of humanoid data covering 260 diverse tasks, ranging from human–object interactions to manipulations of deformable and articulated objects. 
We use a shared action representation for both human hands and robot end-effectors. Specifically, the 48-DoF action in task space is defined as $\mathbf{a}\triangleq\{\mathbf{a}_l, \mathbf{a}_r\}$ and each $\mathbf{a}_l$ or $\mathbf{a}_r \in \mathbb{R}^{24}$ is $\{\mathbf{T}_{wrist}, \mathbf{P}_{thumb}, \mathbf{P}_{index}, \mathbf{P}_{middle}, \mathbf{P}_{ring}, \mathbf{P}_{pinky}\}$. 
The $\mathbf{T} \in \mathbb{R}^9$ is the 9-DoF wrist pose vector consisting of 3D position and 6D rotation. 
Each $\mathbf{P} \in \mathbb{R}^3$ is a 3D fingertip position.
Such unified action representation enables joint training of human and robot data and achieves stable training.

However, naively training the model to autoregressively predict multiple high-dimensional  action chunks is very computationally expensive and drastically slows down pre-training. 
Our key insight is that the goal of pre-training the VLM backbone is to learn the task semantics of the language instruction and the visual representation for downstream real-robot manipulations. 
Predicting a single next-step action suffices for such a goal. 
Therefore, we train the VLM to predict only a single-step action $\mathbf{a}_t$ instead of $\mathbf{a}_{t:t+H}$, which requires much less computation.
We use FAST \cite{pertsch2025fast} to tokenize continuous actions into discrete tokens.
We train the FAST tokenizer on 500,000 randomly sampled actions from EgoDex \cite{egodex}.
The final trained tokenizer achieves an average L1 reconstruction loss of 0.005, and compresses each action sequence from 48 tokens to a variable token length $N\approx20$.
Then, the VLM is trained autoregressively to predict next-action tokens, \textit{i.e.,}  to maximize 
\begin{equation}
p_\theta(\mathbf{a})=\prod^{N}_{t=1}p_\theta(\mathbf{a}_t|\mathbf{a}_{<t}, \ell, \mathbf{o}_t).
\end{equation}

\subsubsection{Post-Training on Cross-Task Real Humanoid Data}
After the VLM backbone is trained, we freeze its parameters and train the action expert from scratch.
Conditioning on the hidden feature extracted from the VLM backbone $\mathbf{z}_t\!=\!f_\theta^{vlm}( \mathbf{o}_t, \ell)$, and a uniformly sampled flow timestep $\tau \in [0,1]$, the flow-matching training objective is 
\begin{equation}
\mathcal{L}_{fm}=\mathbb{E}\left[\lVert v_\rho^{flow}(\mathbf{z}_t, \mathbf{a}_t^\tau, \tau) - (\bm{\epsilon} -\mathbf{a}_t)\rVert\right]
\label{eqn:fm}
\end{equation}
where $\bm{\epsilon}$ is Gaussian noise and
$\mathbf{a}_t^\tau=\tau\mathbf{a}_t+(1-\tau)\bm{\epsilon}$ is the noised action.
We adapt the MM-DiT architecture \cite{sd3} to implement the action expert network $v_\rho^{flow}$, as illustrated in Fig.~\ref{fig:mm-dit}. 
Specifically, the model uses the time-conditioning feature $\tau$ to modulate the action (A) feature and the vision–language (VL) features separately. 
During each transformer block, the action tokens and VL tokens perform joint global attention, which facilitates more effective fusion of visual information compared to the naive DiT.



\subsubsection{Fine-Tuning on In-domain Teleoperation Data}
With the pre-trained VLM and the post-trained action expert, our model can be fine-tuned further end-to-end using a small amount of in-domain data and rapidly learn long-horizon, dexterous loco-manipulation tasks. We evaluate the model on eight real-world tasks (as illustrated in Fig. \ref{fig:eval_tasks}), each posing distinct challenges: some require precise arm coordination, while others demand long-distance navigation. Most tasks exceed 2,000 steps at 30Hz, rendering them truly long-horizon.
Each task contains three to five sub-tasks, and each sub-task corresponds to a skill such as grasping or pushing.

\subsection{Real-Time Action Chunking}

\begin{figure}[t]
  \centering
\includegraphics[width=0.98\columnwidth]{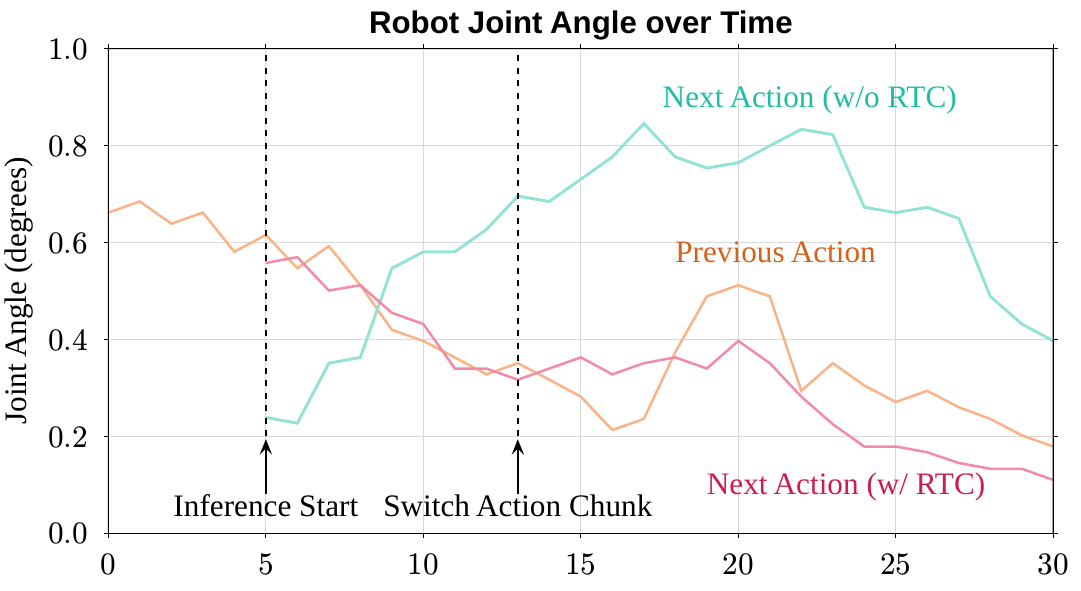}
  \caption{\textbf{Real-Time Chunking:} Given that the previous action is being executed (yellow line), the next action can diverge significantly (cyan line) without RTC, which leads to control jitter. With RTC (red line), the divergence between two consecutive actions is strongly suppressed, resulting in smoother and more stable behavior.}
  \label{fig:rtc-system}
\end{figure}
Humanoid robots require smooth and reactive control, particularly when executing long-horizon, dexterous manipulation tasks. However, existing VLAs typically contain billions of parameters, which inevitably introduce a “stop-and-think” behavior due to inference latency. Our \ours~model similarly comprises over 2.5 billion parameters, with a single forward pass taking approximately 160 ms.
To enable smooth policy rollout despite this latency, we adopt training-time real-time chunking (RTC) following \cite{black2025training}. With RTC, each action prediction is conditioned on the previously committed action chunk and outputs a consistent chunk of future actions, as illustrated in Fig.~\ref{fig:rtc-system}. To faithfully simulate inference delay during training, we randomly remove diffusion noise from the first $d\!=\!\texttt{uniform}(0, d_{\max})$ tokens and mask them out in the loss computation in Eq.~\ref{eqn:fm}. Here, $d_{\max} \in [0, H - s)$ denotes the maximum inference delay in timesteps, while $H$ and $s$ correspond to the action chunk prediction horizon and the execution horizon, respectively.

\subsection{Tailoring Teleoperation for Loco-Manipulation}
\begin{figure}[t]
\centering
\includegraphics[width=0.95\linewidth]{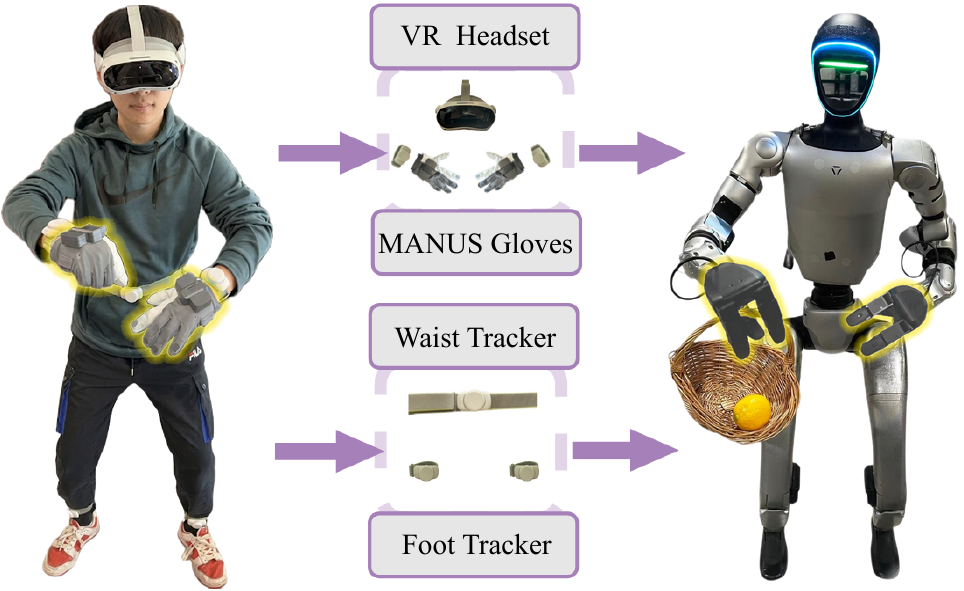}
\caption{\textbf{Real-Robot Teleoperation Setup:} We use MANUS gloves for dexterous hand retargeting; a VR headset and wrist trackers capture upper-body poses for inverse kinematics, while waist and foot trackers provide high-level locomotion commands.}
\label{fig:teleop}
\end{figure}

Efficiently learning a long-horizon loco-manipulation task critically depends on the quality of in-domain data for fine-tuning. 
However, existing teleoperation systems are primarily designed for locomotion and lack the stability and adaptability required for dexterous manipulation.
Designing an effective teleoperation system for humanoid loco-manipulation requires balancing whole-body expressiveness, locomotion stability, and operational simplicity. Existing end-to-end whole-body teleoperation pipelines~\cite{ze2025twist2, luo2025sonic} that directly map full-body human motion to humanoid control through reinforcement learning often suffer from limited robustness due to noisy tracking signals and unstable whole-body motion patterns. Moreover, these systems rely on handheld controllers and reduce dexterous hand control to low-dimensional gripper-like commands, limiting manipulation expressiveness. On the other hand, systems that decouple manipulation from locomotion through explicit base commands~\cite{li2025amo} improve lower-body stability, but typically require additional controllers or multiple operators and thus reduce practicality.

To address these limitations, we propose a tailored teleoperation framework that explicitly separates upper-body pose tracking, dexterous manipulation, and locomotion commands, while enabling single-operator whole-body control. As shown in Fig.~\ref{fig:teleop}, the teleoperator’s upper-body pose is captured using a PICO headset~\cite{pico4ultra} and wrist trackers, and a multi-target inverse kinematics solver is implemented to compute the humanoid’s arm and torso configurations. Fine-grained finger motions are acquired using MANUS gloves~\cite{manusgloves}, allowing direct control over all degrees of freedom of the dexterous hands. Locomotion commands, including translational velocity and turning orientation, are directly inferred from waist and foot trackers and provided as high-level commands to a RL policy~\cite{li2025amo} responsible for stable lower-body control.

By using a small set of wearable trackers and separating locomotion from in-place whole-body actions, our framework enables single-operator humanoid teleoperation with improved locomotion stability across diverse task scenarios. Furthermore, the combination of wrist trackers and MANUS gloves alleviates common occlusion and out-of-view issues in vision-based VR tracking, enabling accurate and reliable upper-body and hand tracking. Together, these design choices support robust and practical humanoid whole-body teleoperation for complex loco-manipulation tasks.

\section{Experiments}
\subsection{Implementation}
\subsubsection{Hardware Platform}
Throughout all real-world experiments, we employ the Unitree G1 humanoid platform, which provides 29 degrees of freedom for whole-body control. In addition, each arm is equipped with a 7-DoF Dex3-1 dexterous hand. Visual observations are obtained using the default head-mounted Intel RealSense D435i camera.

\begin{figure*}[t]
  \centering
  \includegraphics[width=1.0\linewidth]{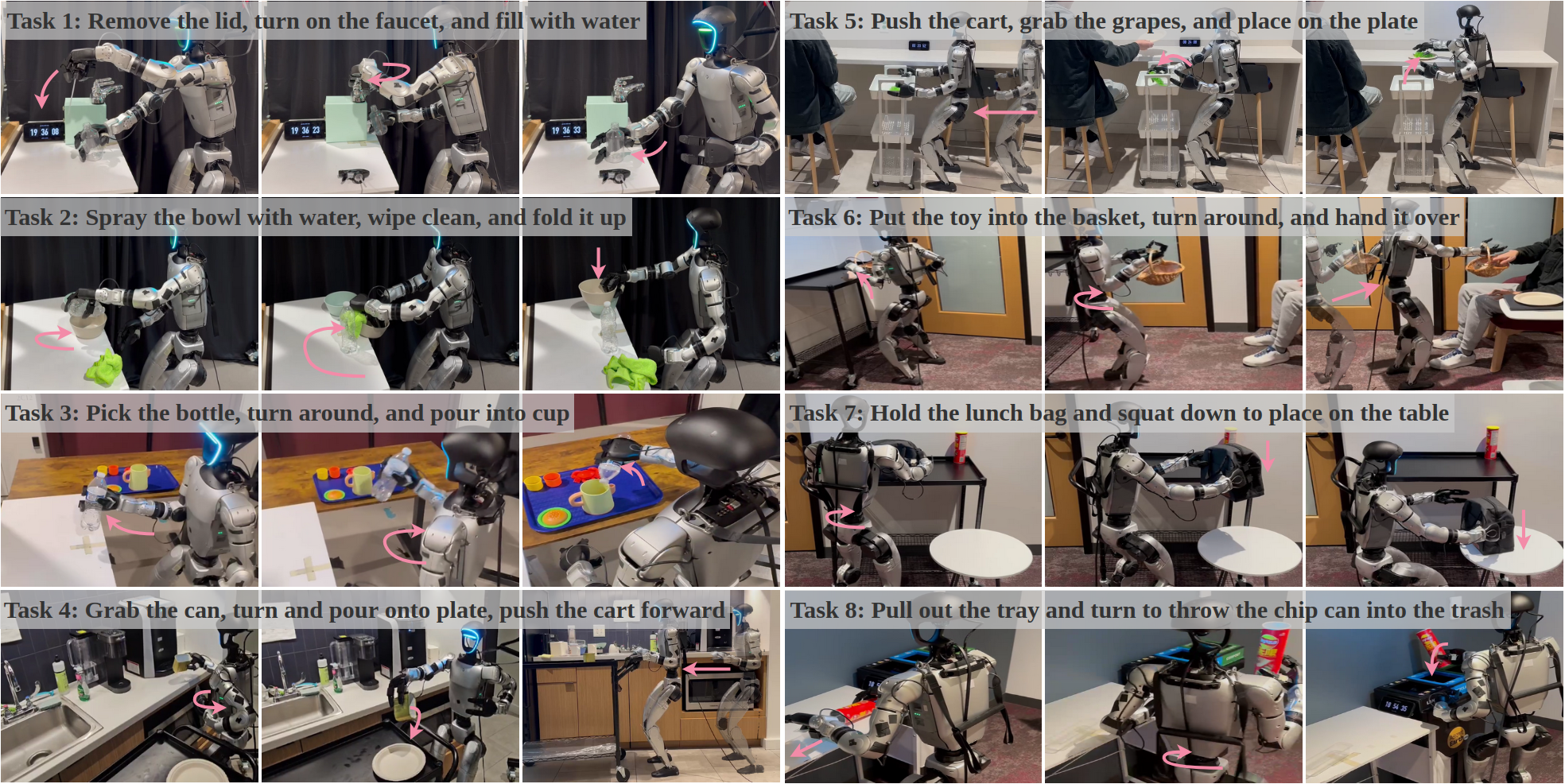}

  \caption{\textbf{Real-World Task Setup:} We evaluate \ours~on eight diverse long-horizon dexterous loco-manipulation tasks involving manipulation, whole-body motion, and locomotion. The task instruction is overlayed on the task images and each sub-task is denoted with marker for better visualization. Our policy rollout videos are included in the Supplementary Materials.}
  \label{fig:eval_tasks} 
\end{figure*}

\subsubsection{Data Preparation}
The EgoDex dataset contains approximately 900M frames and provides per-frame global transformation matrices for the upper humanoid body, including 7 spine joints, 2 arms, and 21 joints for each hand.
To improve pre-training efficiency, all actions are transformed into the current head-camera coordinate frame, and the frame rate is upsampled by a factor of 3.
Due to the presence of extreme outliers in EgoDex, action values are normalized using the 1st and 99th quantiles.
We omit state inputs during the pre-training stage.
We use the Humanoid Everyday dataset \cite{he} for task-agnostic post-training, which contains approximately 3 million frames of real-world teleoperated data.
Actions are represented as 36-DoF joint-space vectors
 $a\!=\!\{\mathbf{q}_{hand}, \mathbf{q}_{arm}, \mathbf{torso}_{rpy}, h_b, v_x, v_y, v_{yaw}, p_{yaw}\}$.
Since Humanoid Everyday only provides upper-body motion, we similarly pad missing lower-body action components.
States consist of 28-DoF joint positions of both hands and arms from the current frame and are fed into the model without normalization.

\subsubsection{Training Details}
Training begins by fitting a FAST tokenizer using 500,000 randomly sampled actions from EgoDex.
The resulting L1 reconstruction loss on held-out action data is approximately 0.005, improving upon the 0.01 using the original open-source FAST tokenizer.
The FAST tokenizer compresses each action sequence into 20 tokens which accelerates subsequent training.
Then, we fine-tune Qwen3-VL-2B-Instruct \cite{Qwen3-VL} during the pre-training stage using 64 A100 GPUs for 10 days. Training is formulated as next-action prediction only, and we avoid action chunking to reduce computational overhead.
The learning rate is fixed at $0.0001$ and the global batch size is 1024.
Next, we post-train the action expert, containing approximately 500M parameters, on the Humanoid Everyday dataset.
During this stage, the VLM backbone is frozen, the learning rate is fixed at 0.0001, and the global batch size is set to 2048.
This stage takes approximately 30 hours on a single node with 32 A100 GPUs.
Finally, we fine-tune only the action expert for each downstream task for 40,000 steps, using a cosine learning rate scheduler with an initial learning rate of 0.0001.


\subsection{Real-World Humanoid Experiments}

\subsubsection{Task Description}
As shown in Fig.~\ref{fig:eval_tasks}, we evaluate \ours~on eight real-world long-horizon manipulation tasks spanning diverse daily scenarios.
The tasks range from simple interactions, such as pick-and-place, pushing, and wiping, to more challenging dexterous manipulations requiring precise finger-object coordination, including turning a faucet and pulling out a chip tray.
Beyond upper-body manipulation, the tasks also involve whole-body motions, such as torso rotation and squatting, as well as lower-body locomotion and turning. 
Overall, this evaluation benchmarks model performance on complex long-horizon dexterous loco-manipulation tasks across multiple real-world environments.

\subsubsection{Evaluation Protocols}
We collect 80 teleoperated trajectories for each task.
All baseline models are fine-tuned on the same dataset, using identical image observations as well as the same action and state representations.
Each long-horizon task consists of three to five sub-tasks involving dexterous manipulation, dual-arm coordination, and locomotion. As a result, policies may fail at early sub-tasks, which can lead to complete rollout failure.
To fully assess the capabilities of each baseline, the evaluator is allowed to intervene and assist the policy in progressing past failed sub-tasks so that execution can continue.
We therefore report success rates for individual sub-tasks in addition to the overall task success rate.
For each task, we perform 10 rollout trials per model.
A rollout is considered successful only if all sub-tasks are completed.
All baselines, including \ours, are deployed using the same client code to control the robot.
\begin{figure*}[t]
  \centering
  \includegraphics[width=0.98\linewidth]{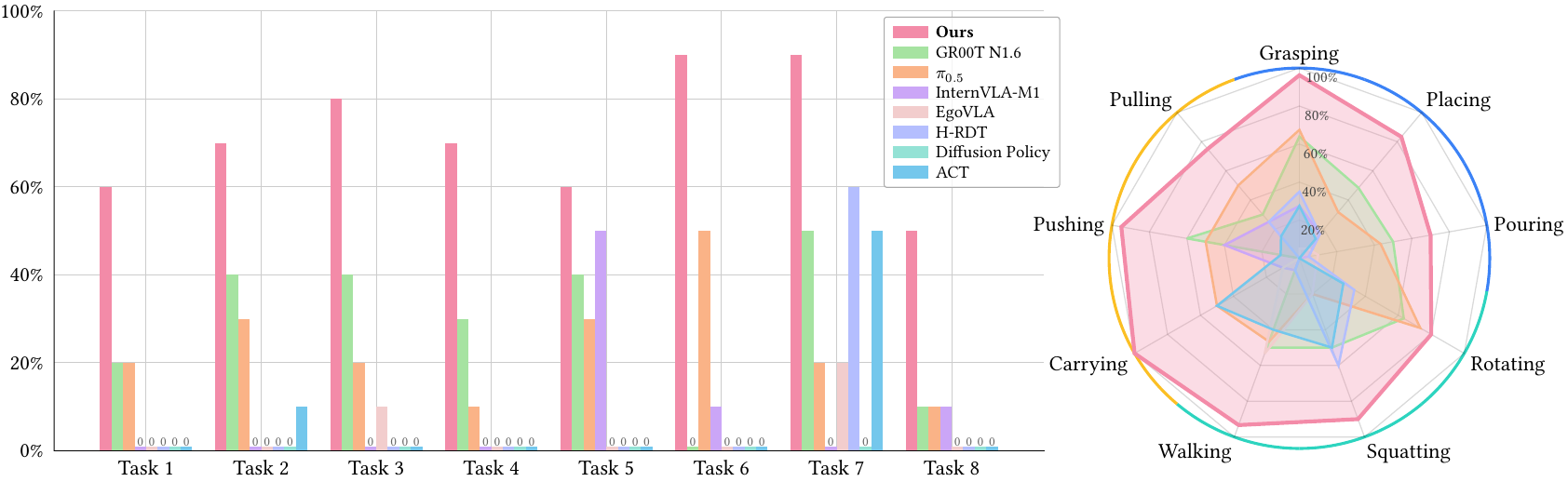}

\caption{
\textbf{Real-World Benchmark:} Evaluation results of policies across our eight tasks, showing task-wise success rates (\%) (left) and aggregated skill-level success rates (\%) (right).
The task descriptions are shown in Fig. \ref{fig:eval_tasks}.
Detailed results for each task including all sub-task progress are included in the \textbf{Supplementary Materials}.
}
  \label{fig:real_world_eval_bar} 
\end{figure*}
\subsubsection{Baselines}
We conduct comprehensive real-world benchmarking against most recent open-source baselines. We invest huge effort to reproduce the best possible results for each. 
\paragraph{$\pi 0.5$} demonstrates strong generalization on mobile robot platforms with dual arms and grippers. 
However, the released model and checkpoint are limited to 30-dimensional action spaces.
To adapt the model to humanoid tasks, we expand the action dimension to 36 and set the action chunk size to 16.
The checkpoint weights of the corresponding linear layers are padded accordingly to accommodate the expanded action space.
To account for the embodiment gap between the original training data and humanoids, we increase the learning rate from 1e-5 to 1e-4 and the global batch size from 32 to 128 for better performance, ensuring fair comparison.
We fine-tune the \textit{Pi05\_DROID} checkpoint, which we convert to a PyTorch implementation.
\paragraph{GR00T N1.6}
shows strong performance in grasping and loco-manipulation, with robust spatial generalization. We use all the default hyperparameters for fine-tuning in the release code.
We initialize the model from the GR00T N1.6 3B pre-trained checkpoint and fine-tune it on our teleoperated data for 20,000 steps with a global batch size of 24 on three NVIDIA A100 GPUs. We use cosine scheduling for the learning rate at 1e-4. As the RTC inference code for GR00T N1.6 is not publicly available in the official repository, we adopt a standard sequential inference scheme, in which the observation corresponding to the most recently executed action is used to condition the prediction of subsequent actions.
\paragraph{InternVLA-M1~\cite{chen2025internvlam1spatiallyguidedvisionlanguageaction}}
is a unified framework for spatial grounding and robot control, which demonstrates strong spatial reasoning capabilities. However, it is only pre-trained on spatial reasoning and robotic arm data which limits its performance on humanoid tasks. We start with the checkpoint pre-trained on the RT-1 Bridge dataset, freeze the VLM backbone and fine-tune the action head for 30,000 steps with a batch size of 64 on a single NVIDIA A100 GPU. 
In our experiments, InternVLA-M1 exhibits action jitter across consecutive action chunks, resulting in unstable executions.

\paragraph{H-RDT}
is a single large DiT action expert with 2B parameters. We train the model for 10,000 training steps with a batch size of 32 on a single NVIDIA A100 GPU. The resulting policy excels at tasks that do not require precise movements. However, it struggles with manipulation tasks that require high-precision across many joints.
\paragraph{EgoVLA}
is a vision–language–action model pre-trained on egocentric human manipulation videos using EgoDex and additional data sources. Since the original codebase predicts only end-effector wrist and hand poses, we adapt the action decoder to output robot joint-space commands required by downstream tasks.
We fine-tune the pre-trained EgoVLA on our teleoperated downstream tasks following the training configuration reported in the original paper, training for 115 epochs with an effective batch size of $16 \times 8 \times 4$. In our experiments, EgoVLA shows limited performance on lower-body commands, likely because its pre-training primarily captures upper-body and hand manipulation skills and does not provide strong priors for coordinated lower-body motion.
\paragraph{Diffusion Policy (DP)~\cite{chi2024diffusionpolicyvisuomotorpolicy}} 

For visual feature extraction, we employ a pre-trained ResNet-18 \cite{7780459} as the visual encoder.
We set the learning rate to $1\times10^{-4}$ and the global batch size to 32. Training is conducted for 40,000 steps using two A100 GPUs, with each task trained for approximately 15 hours.
We observe that DP fails on most tasks, even though it can reasonably fit the training data. 
We conjecture that the UNet-based DP model has insufficient visual capacity.
During inference, we perform 100 iterative denoising steps to progressively transform random noise into actionable trajectories. 

\begin{table*}[h]
\centering
\renewcommand{\arraystretch}{1.3}
\setlength{\tabcolsep}{6pt}
\begin{tabular}{cccccccccc}
\toprule
\multicolumn{2}{c}{Pre-Training} 
& \multirow{2}{*}{\begin{tabular}[c]{@{}c@{}}Post-Training\\ (On HE)\end{tabular}}
& \multirow{2}{*}{\begin{tabular}[c]{@{}c@{}}Real-Time\\ Chunking\end{tabular}}
& \multirow{2}{*}{\begin{tabular}[c]{@{}c@{}} MM-DiT\\ 
Action Head\end{tabular}}
& \multirow{2}{*}{\begin{tabular}[c]{@{}c@{}} Naive DiT\\ 
Action Head\end{tabular}}

& \multirow{2}{*}{\begin{tabular}[c]{@{}c@{}}Right-Arm\\ Pick-n-Place\end{tabular}}
& \multirow{2}{*}{\begin{tabular}[c]{@{}c@{}}Left-Arm\\ Pick-n-Place\end{tabular}}
& \multirow{2}{*}{\begin{tabular}[c]{@{}c@{}}Dual-Arm\\ Carry\end{tabular}}
& \multirow{2}{*}{\begin{tabular}[c]{@{}c@{}}Overall\\ Success Rate\end{tabular}} \\
\cline{1-2}
EgoDex & HE & &&  &  &  &  &  &  \\
\hline
\redx & \redx & \redx & \redx & \redx & \greencheck & 1/10 & 1/10 & 1/10 & 0/10 \\
\redx & \redx & \redx & \redx & \greencheck & \redx & 9/10 & 2/10 & 3/10 & 2/10 \\
\greencheck & \redx & \redx & \redx & \greencheck & \redx & 8/10 & 6/10 & 6/10 & 6/10 \\
\greencheck & \greencheck & \redx & \redx & \greencheck & \redx &  8/10 & 8/10 & 9/10 & 8/10 \\
\greencheck & \greencheck & \greencheck & \redx & \greencheck & \redx & 9/10 & 9/10 & 10/10 & 9/10 \\
\hline
\greencheck & \greencheck & \greencheck & \greencheck & \greencheck & \redx & 9/10 & 9/10 & 9/10 & 9/10 \\
\bottomrule
\end{tabular}
\caption{\textbf{Ablation Studies}. We study the effects of pre-training, post-training, and real-time chunking on a dual-arm long-horizon task which consists of three steps: right-arm pick and place, left-arm pick-and-place and dual-arm lift.}
\label{tab:ablation}
\end{table*}

\paragraph{Action Chunking with Transformers (ACT) \cite{zhao2023learning}} 
To adapt to the  humanoid locomotion and manipulation tasks, we reconfigure the action head to output 36-dimensional actions and tune the chunk size to 100, and initialize the transformer block with a configuration of 4 encoder layers and 1 decoder layer, aligning with the publicly released ACT framework \cite{cadene2024lerobot}.
Other training hyper-parameters like learning rate, batch size and training steps are kept the same as DP.

\subsubsection{Comparisons to Baselines}

As illustrated in Fig. \ref{fig:real_world_eval_bar},  our model outperforms all baselines by a large margin.
Our model exhibits the most stable performance across all eight long-horizon dexterous loco-manipulation tasks.
Notably, it achieves an average overall success rate that is at least 40\% higher than that of the second-best baseline, GR00T-N1.6 \cite{bjorck2025gr00t}, which is the most recently released humanoid foundation model.
These results highlight the effectiveness of our training paradigm, despite using a relatively small amount of robotic data in both the pre-training and post-training stages.
We attribute this success to the unique training recipe.
A key insight is that pre-training the VLM on large-scale human video enables it to learn domain-aligned visual representations for downstream manipulation tasks, while avoiding the hazardous and difficult co-training of two fundamentally different distributions.
With language and visual representations extracted from the pre-trained VLM, we further post-train only the action expert in the joint space using high-quality real-robot data, enabling it to learn a strong prior for embodied control.
More detailed results, including per-subtask progress and policy rollout videos, are provided in the \textbf{Supplemental Material}.


\subsection{Ablation Studies}

Due to limited compute and time, we perform our ablation study using a single real-world task: \textit{pick toys into a box and lift it}.
This task consists of three sub-tasks: (1) picking up a toy dumpling with the right arm and placing it into the box; (2) picking up a toy hippopotamus with the left arm and placing it into the box; and (3) carrying the box with both arms.
This task consists of multiple  execution stages and requires the policy to handle single-arm pick-and-place and dual-arm coordination.
\paragraph{The Role of Pre-Training and Post-Training}
First, we study how the original Qwen3-VL VLM pre-trained on text-generation tasks performs in our settings.
As shown in Table \ref{tab:ablation}, freezing the pre-trained Qwen3-VL backbone and fine-tuning only the action head yields the poorest performance, achieving an overall success rate of only 0.2.
This result highlights the importance of pre-training the VLM backbone on human data to learn how to generate action tokens.
After pre-training on EgoDex for task-space next-action prediction, the model achieves a substantial performance improvement.
Notably, even though the VLM backbone is trained to predict a different action representation than that used by the downstream action head, supervising it with next-step 48-DoF actions still enables the model to learn meaningful visual representations for robotic tasks.
These findings suggest an effective pathway for learning from large-scale human video data while avoiding the inference latency associated with fully autoregressive VLM action generation.
With post-training of the action expert on high-quality robot data, overall performance is further improved.

\paragraph{MM-DiT versus Naive DiT}
We also ablate the effectiveness of the proposed MM-DiT action head by comparing it with a naive DiT for action prediction.
The results show that MM-DiT consistently outperforms the DiT variant.
This improvement can be attributed to MM-DiT’s dual-modulation design and its joint attention mechanism, which integrates VL features from the VLM backbone with Action (A) branch representations. Our analysis suggests that naive DiT, originally designed for text-conditioned image generation, provides weaker conditioning when applied to VL-guided action prediction.
Additional ablation studies on the action expert are provided in the \textbf{Supplementary Material}.

\paragraph{Real-Time Chunking Behaviors}
VLAs typically suffer from slow inference due to their large model size.
When receiving a new query to generate actions, inference can take more than 200 ms, during which the humanoid robot must pause while waiting for the actions to become available, introducing jitter and unstable behavior in whole-body control tasks.
One solution is test-time real-time chunking \cite{black2025real}.
It employs inference-time gradient guidance to the flow-based action generation to steer the future actions to be consistent with past ones, therefore achieving smooth execution of the joint commands. 
However, we found that our model can not be guided at test time stably; as a result,
we implemented training-time real-time chunking \cite{black2025training}.
We observed that real-time chunking mitigates physical collisions during policy execution and increases policy rollout throughput without performance degradation.



\section{Conclusion} 
\label{sec:conclusion}

We introduce \ours\;, an open foundation model accompanied by a complete open-source suite for teleoperation, learning infrastructure, and deployment.
Through extensive experiments, our results suggest that scaling humanoid learning requires scaling the right data in the right way. In contrast to blindly increasing the volume of teleoperation data at substantial cost, we leverage affordable, high-quality  egocentric videos to learn human motion priors and human-object interaction knowledge.  Our work further introduces several novel and empirically validated techniques that significantly improve the effectiveness of humanoid VLAs, including efficient whole-body dexterous teleoperation, MM-DiT-based action experts, and real-time control at deployment. Together, our training recipe and model architecture achieve state-of-the-art performance on challenging, complex, long-horizon tasks, while relying on substantially less real-world robotic data. 
We hope this work can serve as a foundation for humanoid learning, accelerating the development of humanoids capable of assisting with everyday tasks.

\textbf{Limitation.} Due to compute and time constraints, we are unable to further scale training to larger collections of human videos and real-world robotic data, which we leave for future work. Another limitation stems from the hardware platform, whose payload capacity constrains the execution of potentially more capable manipulation behaviors.




\bibliographystyle{plainnat}
\bibliography{references}

@article{li2025amo,
title={AMO: Adaptive Motion Optimization for Hyper-Dexterous Humanoid Whole-Body Control},
author={Li, Jialong and Cheng, Xuxin and Huang, Tianshu and Yang, Shiqi and Qiu, Rizhao and Wang, Xiaolong},
journal={Robotics: Science and Systems 2025},
year={2025}
}

@article{ze2025twist,
title={TWIST: Teleoperated Whole-Body Imitation System},
author= {Yanjie Ze and Zixuan Chen and João Pedro Araújo and Zi-ang Cao and Xue Bin Peng and Jiajun Wu and C. Karen Liu},
year= {2025},
journal= {arXiv preprint arXiv:2505.02833}
}

@misc{li2025clone,
  title={CLONE: Closed-Loop Whole-Body Humanoid Teleoperation for Long-Horizon Tasks}, 
  author={Yixuan Li and Yutang Lin and Jieming Cui and Tengyu Liu and Wei Liang and Yixin Zhu and Siyuan Huang},
  journal={arXiv preprint arXiv:2506.08931}, 
  year={2025}
}

@article{liao2025beyondmimic,
  title={Beyondmimic: From motion tracking to versatile humanoid control via guided diffusion},
  author={Liao, Qiayuan and Truong, Takara E and Huang, Xiaoyu and Gao, Yuman and Tevet, Guy and Sreenath, Koushil and Liu, C Karen},
  journal={arXiv preprint arXiv:2508.08241},
  year={2025}
}

@article{cheng2024expressive,
  title={Expressive whole-body control for humanoid robots},
  author={Cheng, Xuxin and Ji, Yandong and Chen, Junming and Yang, Ruihan and Yang, Ge and Wang, Xiaolong},
  journal={arXiv preprint arXiv:2402.16796},
  year={2024}
}

@article{fu2024humanplus,
  title={Humanplus: Humanoid shadowing and imitation from humans},
  author={Fu, Zipeng and Zhao, Qingqing and Wu, Qi and Wetzstein, Gordon and Finn, Chelsea},
  journal={arXiv preprint arXiv:2406.10454},
  year={2024}
}

@inproceedings{allshire2025visual,
  title={Visual Imitation Enables Contextual Humanoid Control},
  author={Allshire, Arthur and Choi, Hongsuk and Zhang, Junyi and McAllister, David and Zhang, Anthony and Kim, Chung Min and Darrell, Trevor and Abbeel, Pieter and Malik, Jitendra and Kanazawa, Angjoo},
  booktitle={Proceedings of The Conference on Robot Learning},
  series={Proceedings of Machine Learning Research},
  year={2025}
}

@article{ze2025twist2,
  title={Twist2: Scalable, portable, and holistic humanoid data collection system},
  author={Ze, Yanjie and Zhao, Siheng and Wang, Weizhuo and Kanazawa, Angjoo and Duan, Rocky and Abbeel, Pieter and Shi, Guanya and Wu, Jiajun and Liu, C Karen},
  journal={arXiv preprint arXiv:2511.02832},
  year={2025}
}

@article{xue2025leverb,
  title={Leverb: Humanoid whole-body control with latent vision-language instruction},
  author={Xue, Haoru and Huang, Xiaoyu and Niu, Dantong and Liao, Qiayuan and Kragerud, Thomas and Gravdahl, Jan Tommy and Peng, Xue Bin and Shi, Guanya and Darrell, Trevor and Sreenath, Koushil and others},
  journal={arXiv preprint arXiv:2506.13751},
  year={2025}
}

@article{shao2025langwbc,
  title={LangWBC: Language-directed Humanoid Whole-Body Control via End-to-end Learning},
  author={Shao, Yiyang and Huang, Xiaoyu and Zhang, Bike and Liao, Qiayuan and Gao, Yuman and Chi, Yufeng and Li, Zhongyu and Shao, Sophia and Sreenath, Koushil},
  journal={arXiv preprint arXiv:2504.21738},
  year={2025}
}

@inproceedings{grauman2024ego,
  title={Ego-exo4d: Understanding skilled human activity from first-and third-person perspectives},
  author={Grauman, Kristen and Westbury, Andrew and Torresani, Lorenzo and Kitani, Kris and Malik, Jitendra and Afouras, Triantafyllos and Ashutosh, Kumar and Baiyya, Vijay and Bansal, Siddhant and Boote, Bikram and others},
  booktitle={Proceedings of the IEEE/CVF Conference on Computer Vision and Pattern Recognition},
  pages={19383--19400},
  year={2024}
}

@article{luo2025being,
  title={Being-h0: vision-language-action pretraining from large-scale human videos},
  author={Luo, Hao and Feng, Yicheng and Zhang, Wanpeng and Zheng, Sipeng and Wang, Ye and Yuan, Haoqi and Liu, Jiazheng and Xu, Chaoyi and Jin, Qin and Lu, Zongqing},
  journal={arXiv preprint arXiv:2507.15597},
  year={2025}
}

@inproceedings{zitkovich2023rt,
  title={Rt-2: Vision-language-action models transfer web knowledge to robotic control},
  author={Zitkovich, Brianna and Yu, Tianhe and Xu, Sichun and Xu, Peng and Xiao, Ted and Xia, Fei and Wu, Jialin and Wohlhart, Paul and Welker, Stefan and Wahid, Ayzaan and others},
  booktitle={Conference on Robot Learning},
  pages={2165--2183},
  year={2023},
  organization={PMLR}
}

@article{kim2024openvla,
  title={Openvla: An open-source vision-language-action model},
  author={Kim, Moo Jin and Pertsch, Karl and Karamcheti, Siddharth and Xiao, Ted and Balakrishna, Ashwin and Nair, Suraj and Rafailov, Rafael and Foster, Ethan and Lam, Grace and Sanketi, Pannag and others},
  journal={arXiv preprint arXiv:2406.09246},
  year={2024}
}

@article{black2024pi0,
  title        = {{\(\pi_0\)}: A Vision-Language-Action Flow Model for General Robot Control},
  author       = {Black, Kevin and Brown, Noah and Driess, Danny and Esmail, Adnan and Equi, Michael and Finn, Chelsea and Fusai, Niccolo and Groom, Lachy and Hausman, Karol and Ichter, Brian and others},
  journal      = {arXiv preprint arXiv:2410.24164},
  year         = {2024}
}

@article{bjorck2025gr00t,
  title={Gr00t n1: An open foundation model for generalist humanoid robots},
  author={Bjorck, Johan and Casta{\~n}eda, Fernando and Cherniadev, Nikita and Da, Xingye and Ding, Runyu and Fan, Linxi and Fang, Yu and Fox, Dieter and Hu, Fengyuan and Huang, Spencer and others},
  journal={arXiv preprint arXiv:2503.14734},
  year={2025}
}

@article{zhang2024uni,
  title={Uni-navid: A video-based vision-language-action model for unifying embodied navigation tasks},
  author={Zhang, Jiazhao and Wang, Kunyu and Wang, Shaoan and Li, Minghan and Liu, Haoran and Wei, Songlin and Wang, Zhongyuan and Zhang, Zhizheng and Wang, He},
  journal={arXiv preprint arXiv:2412.06224},
  year={2024}
}

@article{geng2023sage,
  title={Sage: Bridging semantic and actionable parts for generalizable articulated-object manipulation under language instructions},
  author={Geng, Haoran and Wei, Songlin and Deng, Congyue and Shen, Bokui and Wang, He and Guibas, Leonidas},
  journal={arXiv preprint arXiv:2312.01307},
  volume={2},
  year={2023}
}

@article{yang2025egovla,
  title={Egovla: Learning vision-language-action models from egocentric human videos},
  author={Yang, Ruihan and Yu, Qinxi and Wu, Yecheng and Yan, Rui and Li, Borui and Cheng, An-Chieh and Zou, Xueyan and Fang, Yunhao and Cheng, Xuxin and Qiu, Ri-Zhao and others},
  journal={arXiv preprint arXiv:2507.12440},
  year={2025}
}

@article{bi2025hrdt,
  title={H-rdt: Human manipulation enhanced bimanual robotic manipulation},
  author={Bi, Hongzhe and Wu, Lingxuan and Lin, Tianwei and Tan, Hengkai and Su, Zhizhong and Su, Hang and Zhu, Jun},
  journal={arXiv preprint arXiv:2507.23523},
  year={2025}
}

@article{cai2025innon,
  title={In-N-On: Scaling Egocentric Manipulation with in-the-wild and on-task Data},
  author={Cai, Xiongyi and Qiu, Ri-Zhao and Chen, Geng and Wei, Lai and Liu, Isabella and Huang, Tianshu and Cheng, Xuxin and Wang, Xiaolong},
  journal={arXiv preprint arXiv:2511.15704},
  year={2025}
}

@article{intelligence2025pi05,
  title        = {{\(\pi_{0.5}\)}: A Vision-Language-Action Model with Open-World Generalization},
  author       = {Physical Intelligence and Black, Kevin and Brown, Noah and Darpinian, James and Dhabalia, Karan and Driess, Danny and Esmail, Adnan and Equi, Michael and Finn, Chelsea and Fusai, Niccolo and others},
  journal      = {arXiv preprint arXiv:2504.16054},
  year         = {2025}
}

@article{team2025gemini,
  title={Gemini robotics 1.5: Pushing the frontier of generalist robots with advanced embodied reasoning, thinking, and motion transfer},
  author={Team, Gemini Robotics and Abdolmaleki, Abbas and Abeyruwan, Saminda and Ainslie, Joshua and Alayrac, Jean-Baptiste and Arenas, Montserrat Gonzalez and Balakrishna, Ashwin and Batchelor, Nathan and Bewley, Alex and Bingham, Jeff and others},
  journal={arXiv preprint arXiv:2510.03342},
  year={2025}
}

@article{Qwen3-VL,
      title={Qwen3-VL Technical Report}, 
      author={Shuai Bai and Yuxuan Cai and Ruizhe Chen and Keqin Chen and Xionghui Chen and Zesen Cheng and Lianghao Deng and Wei Ding and Chang Gao and Chunjiang Ge and Wenbin Ge and Zhifang Guo and Qidong Huang and Jie Huang and Fei Huang and Binyuan Hui and Shutong Jiang and Zhaohai Li and Mingsheng Li and Mei Li and Kaixin Li and Zicheng Lin and Junyang Lin and Xuejing Liu and Jiawei Liu and Chenglong Liu and Yang Liu and Dayiheng Liu and Shixuan Liu and Dunjie Lu and Ruilin Luo and Chenxu Lv and Rui Men and Lingchen Meng and Xuancheng Ren and Xingzhang Ren and Sibo Song and Yuchong Sun and Jun Tang and Jianhong Tu and Jianqiang Wan and Peng Wang and Pengfei Wang and Qiuyue Wang and Yuxuan Wang and Tianbao Xie and Yiheng Xu and Haiyang Xu and Jin Xu and Zhibo Yang and Mingkun Yang and Jianxin Yang and An Yang and Bowen Yu and Fei Zhang and Hang Zhang and Xi Zhang and Bo Zheng and Humen Zhong and Jingren Zhou and Fan Zhou and Jing Zhou and Yuanzhi Zhu and Ke Zhu},
	  journal={arXiv preprint arXiv:2511.21631},
      year={2025}
}

@inproceedings{sd3,
  title={Scaling rectified flow transformers for high-resolution image synthesis},
  author={Esser, Patrick and Kulal, Sumith and Blattmann, Andreas and Entezari, Rahim and M{\"u}ller, Jonas and Saini, Harry and Levi, Yam and Lorenz, Dominik and Sauer, Axel and Boesel, Frederic and others},
  booktitle={Forty-first international conference on machine learning},
  year={2024}
}

@misc{egodex,
      title={EgoDex: Learning Dexterous Manipulation from Large-Scale Egocentric Video}, 
      author={Ryan Hoque and Peide Huang and David J. Yoon and Mouli Sivapurapu and Jian Zhang},
      year={2025},
      eprint={2505.11709},
      archivePrefix={arXiv},
      primaryClass={cs.CV},
      url={https://arxiv.org/abs/2505.11709}, 
}

@article{pertsch2025fast,
  title={Fast: Efficient action tokenization for vision-language-action models},
  author={Pertsch, Karl and Stachowicz, Kyle and Ichter, Brian and Driess, Danny and Nair, Suraj and Vuong, Quan and Mees, Oier and Finn, Chelsea and Levine, Sergey},
  journal={arXiv preprint arXiv:2501.09747},
  year={2025}
}

@article{he,
  title={Humanoid everyday: A comprehensive robotic dataset for open-world humanoid manipulation},
  author={Zhao, Zhenyu and Jing, Hongyi and Liu, Xiawei and Mao, Jiageng and Jha, Abha and Yang, Hanwen and Xue, Rong and Zakharor, Sergey and Guizilini, Vitor and Wang, Yue},
  journal={arXiv preprint arXiv:2510.08807},
  year={2025}
}

@inproceedings{uh1,
  title={Universal humanoid robot pose learning from internet human videos},
  author={Mao, Jiageng and Zhao, Siheng and Song, Siqi and Hong, Chuye and Shi, Tianheng and Ye, Junjie and Zhang, Mingtong and Geng, Haoran and Malik, Jitendra and Guizilini, Vitor and others},
  booktitle={2025 IEEE-RAS 24th International Conference on Humanoid Robots (Humanoids)},
  pages={1--8},
  year={2025},
  organization={IEEE}
}

@article{luo2025sonic,
  title={Sonic: Supersizing motion tracking for natural humanoid whole-body control},
  author={Luo, Zhengyi and Yuan, Ye and Wang, Tingwu and Li, Chenran and Chen, Sirui and Casta{\~n}eda, Fernando and Cao, Zi-Ang and Li, Jiefeng and Minor, David and Ben, Qingwei and others},
  journal={arXiv preprint arXiv:2511.07820},
  year={2025}
}

@misc{pico4ultra,
  author       = {{PICO Immersive Pte. Ltd.}},
  title        = {{PICO 4 Ultra: An All-New Mixed Reality Experience}},
  year         = {2023},
  howpublished = {\url{https://www.picoxr.com/global/products/pico4-ultra}},
}

@misc{manusgloves,
  author       = {{MANUS Technology Group}},
  title        = {{MANUS – High-Precision Data Gloves for Robotics, VR \& Mocap}},
  year         = {2024},
  howpublished = {\url{https://www.manus-meta.com/}},
}

@article{black2025real,
  title={Real-Time Execution of Action Chunking Flow Policies},
  author={Black, Kevin and Galliker, Manuel Y and Levine, Sergey},
  journal={arXiv preprint arXiv:2506.07339},
  year={2025}
}

@article{zhao2023learning,
  title={Learning fine-grained bimanual manipulation with low-cost hardware},
  author={Zhao, Tony Z and Kumar, Vikash and Levine, Sergey and Finn, Chelsea},
  journal={arXiv preprint arXiv:2304.13705},
  year={2023}
}

@article{liu2024bidirectional,
  title={Bidirectional decoding: Improving action chunking via closed-loop resampling},
  author={Liu, Yuejiang and Hamid, Jubayer Ibn and Xie, Annie and Lee, Yoonho and Du, Maximilian and Finn, Chelsea},
  journal={arXiv preprint arXiv:2408.17355},
  year={2024}
}

@article{black2025training,
  title={Training-time action conditioning for efficient real-time chunking},
  author={Black, Kevin and Ren, Allen Z and Equi, Michael and Levine, Sergey},
  journal={arXiv preprint arXiv:2512.05964},
  year={2025}
}

@misc{chi2024diffusionpolicyvisuomotorpolicy,
      title={Diffusion Policy: Visuomotor Policy Learning via Action Diffusion}, 
      author={Cheng Chi and Zhenjia Xu and Siyuan Feng and Eric Cousineau and Yilun Du and Benjamin Burchfiel and Russ Tedrake and Shuran Song},
      year={2024},
      eprint={2303.04137},
      archivePrefix={arXiv},
      primaryClass={cs.RO},
      url={https://arxiv.org/abs/2303.04137}, 
}

@article{brohan2022rt,
  title={Rt-1: Robotics transformer for real-world control at scale},
  author={Brohan, Anthony and Brown, Noah and Carbajal, Justice and Chebotar, Yevgen and Dabis, Joseph and Finn, Chelsea and Gopalakrishnan, Keerthana and Hausman, Karol and Herzog, Alex and Hsu, Jasmine and others},
  journal={arXiv preprint arXiv:2212.06817},
  year={2022}
}

@article{liu2024visual,
  title={Visual whole-body control for legged loco-manipulation},
  author={Liu, Minghuan and Chen, Zixuan and Cheng, Xuxin and Ji, Yandong and Qiu, Ri-Zhao and Yang, Ruihan and Wang, Xiaolong},
  journal={arXiv preprint arXiv:2403.16967},
  year={2024}
}

@article{ding2025humanoid,
  title={Humanoid-vla: Towards universal humanoid control with visual integration},
  author={Ding, Pengxiang and Ma, Jianfei and Tong, Xinyang and Zou, Binghong and Luo, Xinxin and Fan, Yiguo and Wang, Ting and Lu, Hongchao and Mo, Panzhong and Liu, Jinxin and others},
  journal={arXiv preprint arXiv:2502.14795},
  year={2025}
}

@article{jiang2025wholebodyvla,
  title={WholeBodyVLA: Towards Unified Latent VLA for Whole-Body Loco-Manipulation Control},
  author={Jiang, Haoran and Chen, Jin and Bu, Qingwen and Chen, Li and Shi, Modi and Zhang, Yanjie and Li, Delong and Suo, Chuanzhe and Wang, Chuang and Peng, Zhihui and others},
  journal={arXiv preprint arXiv:2512.11047},
  year={2025}
}

@misc{cadene2024lerobot,
    author = {Cadene, Remi and Alibert, Simon and Soare, Alexander and Gallouedec, Quentin and Zouitine, Adil and Palma, Steven and Kooijmans, Pepijn and Aractingi, Michel and Shukor, Mustafa and Aubakirova, Dana and Russi, Martino and Capuano, Francesco and Pascal, Caroline and Choghari, Jade and Moss, Jess and Wolf, Thomas},
    title = {LeRobot: State-of-the-art Machine Learning for Real-World Robotics in Pytorch},
    howpublished = "\url{https://github.com/huggingface/lerobot}",
    year = {2024}
}

@INPROCEEDINGS{7780459,
  author={He, Kaiming and Zhang, Xiangyu and Ren, Shaoqing and Sun, Jian},
  booktitle={2016 IEEE Conference on Computer Vision and Pattern Recognition (CVPR)}, 
  title={Deep Residual Learning for Image Recognition}, 
  year={2016},
  volume={},
  number={},
  pages={770-778},
  doi={10.1109/CVPR.2016.90}
}

@misc{kareer2024egomimicscalingimitationlearning,
  title={EgoMimic: Scaling Imitation Learning via Egocentric Video}, 
  author={Simar Kareer and Dhruv Patel and Ryan Punamiya and Pranay Mathur and Shuo Cheng and Chen Wang and Judy Hoffman and Danfei Xu},
  year={2024},
  eprint={2410.24221},
  archivePrefix={arXiv},
  primaryClass={cs.RO},
  url={https://arxiv.org/abs/2410.24221},
}

@misc{zhao2025resmimicgeneralmotiontracking,
    title={ResMimic: From General Motion Tracking to Humanoid Whole-body Loco-Manipulation via Residual Learning}, 
    author={Siheng Zhao and Yanjie Ze and Yue Wang and C. Karen Liu and Pieter Abbeel and Guanya Shi and Rocky Duan},
    year={2025},
    eprint={2510.05070},
    archivePrefix={arXiv},
    primaryClass={cs.RO},
    url={https://arxiv.org/abs/2510.05070}, 
}

@article{qi2025coordinated,
 title={Coordinated Humanoid Manipulation with Choice Policies},
 author={Qi, Haozhi and Wang, Yen-Jen and Lin, Toru and Yi Brent and Ma, Yi and Sreenath, Koushil and Malik, Jitendra},
 journal={arXiv:2512.25072},
 year={2025}
}

@misc{yu2025egomilearningactivevision,
  title={EgoMI: Learning Active Vision and Whole-Body Manipulation from Egocentric Human Demonstrations}, 
  author={Justin Yu and Yide Shentu and Di Wu and Pieter Abbeel and Ken Goldberg and Philipp Wu},
  year={2025},
  eprint={2511.00153},
  archivePrefix={arXiv},
  primaryClass={cs.RO},
  url={https://arxiv.org/abs/2511.00153}, 
}

@misc{chen2025internvlam1spatiallyguidedvisionlanguageaction,
      title={InternVLA-M1: A Spatially Guided Vision-Language-Action Framework for Generalist Robot Policy}, 
      author={Xinyi Chen and Yilun Chen and Yanwei Fu and Ning Gao and Jiaya Jia and Weiyang Jin and Hao Li and Yao Mu and Jiangmiao Pang and Yu Qiao and Yang Tian and Bin Wang and Bolun Wang and Fangjing Wang and Hanqing Wang and Tai Wang and Ziqin Wang and Xueyuan Wei and Chao Wu and Shuai Yang and Jinhui Ye and Junqiu Yu and Jia Zeng and Jingjing Zhang and Jinyu Zhang and Shi Zhang and Feng Zheng and Bowen Zhou and Yangkun Zhu},
      year={2025},
      eprint={2510.13778},
      archivePrefix={arXiv},
      primaryClass={cs.RO},
      url={https://arxiv.org/abs/2510.13778}, 
}

\clearpage
\tableofcontents

\section{More Training Details}
\subsection{Pre-Training}
\textbf{FAST Tokenization:} We use the data processing script from H-RDT \cite{bi2025hrdt} to obtain a 48-DoF task-space action representation, along with the corresponding dataset statistics.
The action data is down-sampled from the original 30 Hz to 10 Hz.
We find that the original open-sourced FAST tokenizer \cite{pertsch2025fast} exhibits a large reconstruction loss ($0.583 \times 10^{-4}$), particularly under noisy token settings.
To address this issue, we trained the FAST tokenizer from scratch using 500,000 randomly sampled actions, leading to longer token lengths.
Actions are normalized using the 1st and 99th quantiles of the dataset.
The action horizon, vocabulary size, and scale are set to 1, 2048, and 100, respectively.
A comparison of action reconstruction performance before and after fitting is shown in Table~\ref{tab:fast}.

\begin{table}[h]
\centering
\begin{tabularx}{\linewidth}{p{1.5cm}cc}
\toprule
& Reconstruction L1 Loss & Avg Token Length\\
\midrule
Before & $5.83\times1\mathrm{e}{-}4$ & \textbf{2.08}\\
After& $\textbf{1.95}\times1\mathrm{e}{-}4 $ & 13.04 \\
\bottomrule
\end{tabularx}
\caption{\textbf{Fast Tokenizer.} Comparison of reconstruction loss and average token length before and after training. Boldface indicates the best performance.}
\label{tab:fast}
\end{table}

\textbf{Hyper-Parameters:}
We train the full VLM backbone using DeepSpeed, following the original Qwen3-VL training setup \cite{Qwen3-VL}.
The learning rates for the language backbone, MM projector, and vision tower are set to $1 \times 10^{-4}$, $1 \times 10^{-5}$, and $1 \times 10^{-5}$, respectively, and are kept constant throughout pre-training.
We observe that the default learning rate of $1 \times 10^{-6}$ is too small for effective convergence.
The default image resolution in EgoDex is $1920 \times 1080$, which is prohibitively memory-intensive; therefore, we resize images to $360 \times 240$.
We pre-train the Qwen3-VL-2B-Instruct variant using 64 A100 GPUs with a global batch size of 1024.
Training takes approximately 10 days to reach 230k steps, where the first 200k steps are trained exclusively on the EgoDex dataset and the remaining 30k steps are trained solely on the Humanoid Everyday dataset \cite{he}.

\subsection{Post-Training}
\textbf{Data Processing:}
We post-train the action expert in joint space using the Humanoid Everyday (HE) dataset \cite{he}.
Since HE contains two different embodiments—G1 with Dex3-1 and H1 with the Inspire Hand—which have different finger joint morphologies and degrees of freedom, we align the action representations by reordering the default joint indices.
The resulting action representation has 28 DoF, consisting of 14 DoF for the hand and 14 DoF for the arm.
The state representation is processed in a similar manner.
To enable future fine-tuning of the action expert without reinitializing the state and action projectors, we pad the action and state vectors to 36 DoF and 32 DoF, respectively.
The padded dimensions correspond to lower-body control signals that are not present in the HE dataset.

\textbf{Hyper-Parameters:}
During post-training, the VLM backbone is frozen, and only the action expert is optimized using a constant learning rate of $1 \times 10^{-4}$.
The global batch size is set to 2048, and training is conducted for 30k steps.
Training took approximately 30 hours on 32 A100 GPUs.
Input images are down-scaled to $320 \times 240$.
We adopt uniform sampling for the diffusion time steps $\tau \in [0,1]$ and observe no performance difference compared to alternative sampling strategies \cite{intelligence2025pi05,bjorck2025gr00t} in our real-world experiments.

\begin{figure}[h]
\centering
\includegraphics[width=0.95\columnwidth]{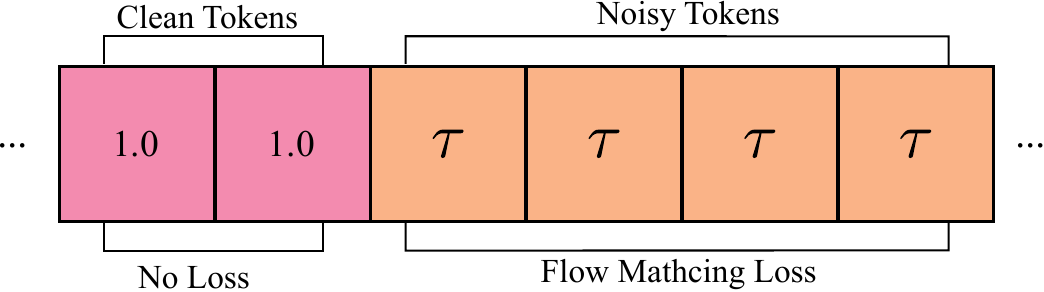}
\caption{\textbf{Training-Time RTC.} Diffusion timesteps and loss calculation in training.}
\label{fig:rtc}
\end{figure}

\subsection{Fine-Tuning}
For real-world tasks, we fine-tune only the action expert while keeping the VLM backbone frozen.
Each real-world task consists of 80 episodes of teleoperation data.
We set the global batch size to 128 and train for 40k steps per task.
A cosine learning rate scheduler is used, with the initial learning rate set to $1 \times 10^{-4}$.
The state and action are normalized using their respective minimum and maximum values.
The image resolution and diffusion timestep sampling follow the same settings as in post-training.
Support for real-time chunking is described in Section~\ref{ssec:rtc-training}.

\begin{figure*}[t]
\centering
\includegraphics[width=0.98\textwidth]{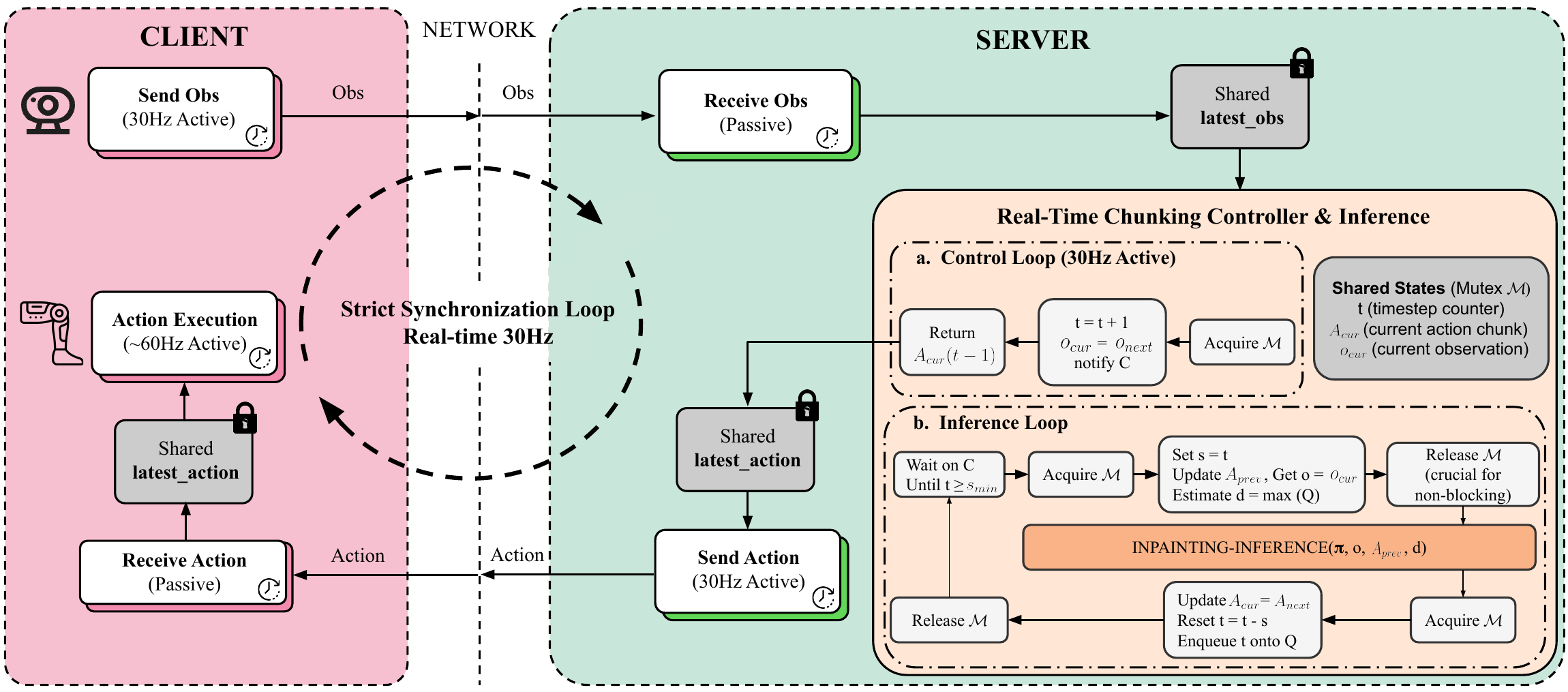}

\caption{\textbf{Real-Time Action Chunking System Design.} 
The system consists of a client (observation collection and action 
execution) and a server (control and inference). The Control Loop 
(30Hz) coordinates observation updates and action dispatch, while 
the Inference Loop runs asynchronously to compute the next action 
chunk when $t \geq s_{\text{min}}$, enabling seamless chunk 
transitions without inference-induced interruptions.}
\label{fig:rtc-figure} 
\end{figure*}

\section{Real-Time Chunking}
\subsection{Training-Time RTC}
\label{ssec:rtc-training}
In addition to training, \ours\;enables real-time control at deployment time. Modern VLAs usually have billions of parameters \cite{kim2024openvla,team2025gemini,black2025real}, leading to substantial inference latency with naive synchronous inference strategies. Specifically, with the naive "stop-think-execute" strategy, rollouts exhibit visible pauses and even jitters between consecutive action chunks. Introducing pauses between chunks not only slows down the rollout process but also creates a training-evaluation gap, which will cause a higher failure rate \cite{black2025real}.

A straightforward approach to address this issue is naive action chunking, which starts the next inference before the previous action chunk is fully executed and switches to the new chunk once it becomes available. While this strategy mitigates the inference delay problem, it introduces jittery transitions between chunks due to randomness and discontinuity, which can be even more detrimental to rollout performance. To address this limitation, recent work has explored methods to maintain continuity between chunks \cite{zhao2023learning, liu2024bidirectional,black2025real,black2025training}. Among these approaches, real-time chunking with training-time~\cite{black2025training} or test-time~\cite{black2025real} action inpainting method demonstrates the best performance.

In practice, we found that our model cannot be guided stably at test time~\cite{black2025real}; as a result, we implemented
training-time real-time chunking~\cite{black2025training}. Unlike test-time RTC, which only requires correcting the velocity $v$ (in flow matching) or noise $\epsilon$ (in diffusion models) predicted by the action head during inference, training-time RTC necessitates modifying the model during the training phase. Specifically, we randomly mask the first $d \in [1, d_{\max}]$ action tokens, where $d_{\max}$ is set to 6 in our experiments.
The masked action tokens are excluded from loss computation, as illustrated in Fig.~\ref{fig:rtc}.
The model is trained to predict actions conditioned on the preceding \textit{clean} action tokens, so that it can generate the remaining tokens with smooth continuity to the \textit{clean} action tokens.
During inference, action steps that have not yet been executed are treated as clean tokens and are used to generate the next action chunk.

\subsection{System Implementation}
We demonstrate our real-time action chunking system design in Fig. \ref{fig:rtc-figure}. The system consists of two components: a client for obtaining observations and executing actions, and a server for control and model inference. The overall operating frequency is determined by the Control Loop on the server side, running at 30Hz. At each timestep in the Control Loop, the observation is updated, an action is queried and sent to the client for execution, which then generates a new observation.

To ensure uninterrupted action execution, model inference runs asynchronously with action execution, controlled by the Inference Loop. The Inference Loop shares the action chunk, observation, and timestep counter with the Control Loop. When the current action chunk has been executed beyond a certain threshold ($t \geq s_{\text{min}}$), the inference loop is triggered to obtain the next action chunk. The system switches to the new action chunk before the previous one completes, ensuring that no system interruption occurs between action chunks due to inference latency.

\section{Whole-Body Teleoperation Pipeline}
\subsection{Whole-Body Control}
\begin{figure}[h]
\centering
\includegraphics[width=0.95\columnwidth]{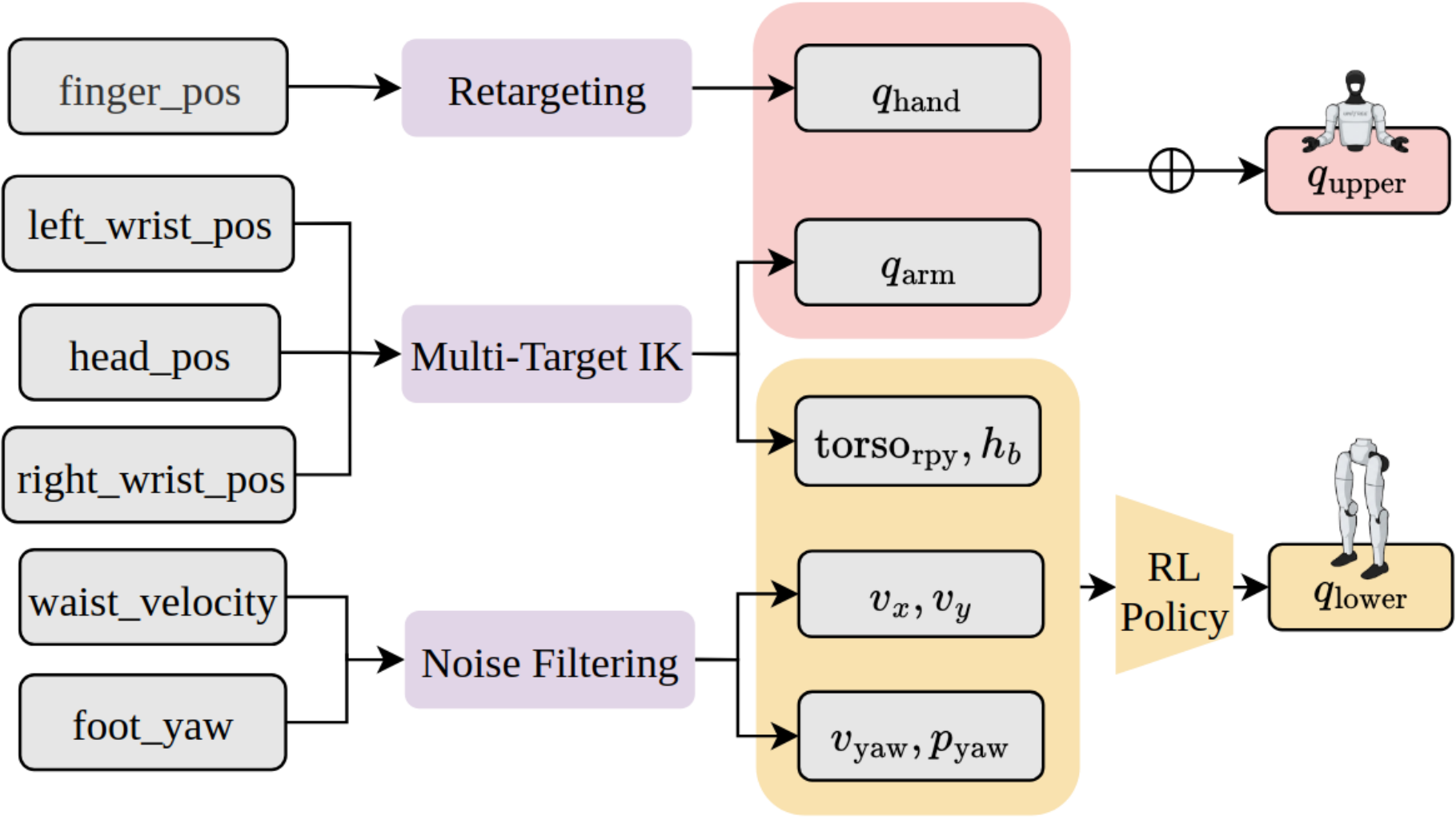}
\caption{\textbf{Single Operator Teleoperation Framework.} Our framework maps human upper-body motions to robot arm and hand control via retargeting and multi-target IK, while lower-body pose is generated through an RL-based policy.}
\label{fig:teleoperation}
\end{figure}

As shown in Fig.~\ref{fig:teleoperation}, using the PICO4U~\cite{pico4ultra} headset together with two wrist trackers, we treat the head and wrist poses as three end-effectors and solve a multi-target inverse kinematics (IK) problem.
This directly produces the humanoid arm joint positions $q_{\text{arm}}$, as well as intermediate variables including torso orientation $\text{torso}_{\text{rpy}}$ and pelvis height $h_b$, which modulate the robot's upper-body posture. These intermediate variables are further provided to a low-level locomotion RL policy (AMO)~\cite{li2025amo}, which outputs the lower-body joint states $q_{\text{lower}}$.

This hierarchical design enables coordinated whole-body control while maintaining balance and locomotion stability.

\subsection{Dexterous Manipulation}
We use MANUS gloves~\cite{manusgloves} to obtain accurate finger tracking from the teleoperator. The thumb, index finger, and middle finger motions are retargeted to the three-finger dexterous hands of the G1 humanoid to enable dexterous manipulation. By combining MANUS gloves with PICO wrist trackers, we directly obtain reliable hand and wrist end-effector poses without relying on unstable vision-based VR hand tracking. This design avoids common occlusion and out-of-view issues and provides more precise hand pose estimation for whole-body dexterous manipulation.


\subsection{Locomotion}
Unlike prior approaches such as TWIST2~\cite{ze2025twist2} and SONIC~\cite{luo2025sonic}, we do not directly retarget the whole-body SMPL motion provided by the PICO tracking system to the humanoid. We find that end-to-end whole-body tracking and retargeting is often not robust, frequently leading to foot drifting, unstable lower body motion, and excessive small corrective steps that hinder policy learning. Instead, we control locomotion through high-level commands $(v_x, v_y, v_{\text{yaw}}, p_{\text{yaw}})$. The PICO waist tracker estimates the operator's translational velocity $(v_x, v_y)$, which is mapped to the robot's base translation. In addition, the foot trackers provide signals to compute yaw commands ($v_{\text{yaw}}$, $p_{\text{yaw}}$) for regulating the humanoid’s base orientation. We also apply clipping and filtering to suppress noise caused by natural human body sway, ensuring accurate locomotion command estimation.

\vspace{0.5em}

Overall, our teleoperation pipeline enables a single operator to perform stable humanoid whole-body teleoperation and execute complex dexterous loco-manipulation tasks.

\section{Real-World Experiment Details}
\subsection{Task Descriptions}
We evaluate \ours~on eight real-world long-horizon manipulation tasks spanning diverse daily scenarios.
\textit{\textcolor[HTML]{fa670c}{The policy rollouts for all tasks are included in the supplementary video.}}


\textit{\textbf{Task 1. Remove the lid, turn on the faucet, and fill with water:}}
The robot grasps the spray bottle with its right hand, and removes the lid from the bottle with left hand and places it on the table.
The robot then moves the bottle under the faucet. With the index finger of its left hand, the robot turns the faucet clockwise to start the water flow and fills the bottle with water. Finally, the robot places the filled bottle back on the table.

\begin{table*}[t]

\centering
\footnotesize 
\setlength{\tabcolsep}{3pt} 

\begin{tabular*}{\textwidth}{@{\extracolsep{\fill}} l | ccccc r | ccccc r @{}}
\toprule
\multicolumn{13}{l}{\textbf{Task 3} \hfill \textbf{Task 2}} \\
\midrule
Descriptions
& \multicolumn{6}{l|}{\makecell[l]{Pick the bottle, turn around, and pour into cup}}
& \multicolumn{6}{l}{\makecell[l]{Spray the bowl with water,\\ wipe clean, and fold
it up}} \\
\midrule
& Grasp & Move & Pour & Place & & \textbf{Overall}
& Grasp & Pull & Spray & Put & & \textbf{Overall} \\
\midrule
Diffusion Policy & 0/10 & 0/10 & 0/10 & 0/10 & & 0/10 & 0/10 & 0/10 & 0/10 & 0/10 & & 0/10 \\
ACT              & 0/10 & 0/10 & 0/10 & 0/10 & & 0/10 & 3/10 & 2/10 & 4/10 & 3/10 & & 1/10 \\
InternVLA-M1     & 0/10 & 0/10 & 0/10 & 0/10 & & 0/10 & 0/10 & 0/10 & 2/10 & \underline{4/10} & & 0/10 \\
EgoVLA           & 4/10 & \underline{6/10} & 1/10 & 2/10 & & 1/10 & 0/10 & 0/10 & 0/10 & 0/10 & & 0/10 \\
H-RDT            & 1/10 & 0/10 & 0/10 & 1/10 & & 0/10 & 0/10 & 1/10 & 0/10 & 0/10 & & 0/10 \\
$\pi 0.5$        & \textbf{10/10} & \underline{6/10} & 3/10 & 2/10 & & 2/10 & \underline{9/10} & \underline{7/10} & \underline{5/10} & \textbf{7/10} & & 3/10 \\
GR00T N1.6       & 3/10 & 5/10 & \underline{5/10} & \underline{4/10} & & \underline{4/10} & 5/10 & 5/10 & \textbf{9/10} & \textbf{7/10} & & \underline{4/10} \\
\textbf{\ours (Ours)}    & \underline{9/10} & \textbf{8/10} & \textbf{8/10} & \textbf{8/10} & & \textbf{8/10} & \textbf{10/10} & \textbf{10/10} & \textbf{9/10} & \textbf{7/10} & & \textbf{7/10} \\

\midrule
\multicolumn{13}{l}{\textbf{Task 4} \hfill \textbf{Task 1}} \\
\midrule
Descriptions
& \multicolumn{6}{l|}{\makecell[l]{Grab the can, turn and pour onto plate, \\ push
the cart forward}}
& \multicolumn{6}{l}{\makecell[l]{Remove the lid, turn on the faucet,\\ and fill with
water}} \\
\midrule
& Grasp & Rotate & Pour & Grab & Push & \textbf{Overall}
& Grasp & Remove & Turn & Put & & \textbf{Overall} \\
\midrule
Diffusion Policy & 0/10 & 0/10 & 0/10 & 0/10 & 0/10 & 0/10 & 0/10 & 0/10 & 0/10 & 0/10 & & 0/10 \\
ACT              & 0/10 & 0/10 & 0/10 & 0/10 & 0/10 & 0/10 & \underline{7/10} & 0/10 & 0/10 & 0/10 & & 0/10 \\
InternVLA-M1     & 2/10 & 0/10 & 1/10 & 0/10 & 0/10 & 0/10 & 0/10 & 0/10 & 0/10 & 0/10 & & 0/10 \\
EgoVLA           & 0/10 & 0/10 & 1/10 & 0/10 & 0/10 & 0/10 & 0/10 & 0/10 & 0/10 & 0/10 & & 0/10 \\
H-RDT            & 3/10 & 1/10 & 1/10 & 0/10 & 0/10 & 0/10 & \underline{7/10} & 0/10 & 0/10 & 1/10 & & 0/10 \\
$\pi 0.5$        & 2/10 & 5/10 & 5/10 & 8/10 & 1/10 & 1/10 & 4/10 & \underline{4/10} & \underline{8/10} & 2/10 & & \underline{2/10} \\
GR00T N1.6       & \underline{5/10} & \underline{7/10} & \underline{5/10} & \underline{4/10} & \underline{3/10} & \underline{3/10} & \textbf{10/10} & 3/10 & 2/10 & \underline{3/10} & & \underline{2/10} \\
\textbf{\ours (Ours)}    & \textbf{10/10} & \textbf{9/10} & \textbf{7/10} & \textbf{10/10} & \textbf{10/10} & \textbf{7/10} & \textbf{10/10} & \textbf{10/10} & \textbf{6/10} & \textbf{10/10} & & \textbf{6/10} \\

\midrule
\multicolumn{13}{l}{\textbf{Task 5} \hfill \textbf{Task 6}} \\
\midrule
Descriptions
& \multicolumn{6}{l|}{\makecell[l]{Put toy into basket, walk\\ to human, hand it over}}
& \multicolumn{6}{l}{\makecell[l]{Push the cart, grab the grapes,\\ and place on the
plate}} \\
\midrule
& Grasp & Hook & Walk & Hand & & \textbf{Overall}
& Handle & Push & Grasp & Place & & \textbf{Overall} \\
\midrule
Diffusion Policy & 0/10 & 0/10 & 0/10 & 0/10 & & 0/10 & 0/10 & 0/10 & 0/10 & 0/10 & & 0/10 \\
ACT              & 3/10 & 0/10 & 5/10 & \underline{5/10} & & 0/10 & 2/10 & 2/10 & 0/10 & 0/10 & & 0/10 \\
InternVLA-M1     & 2/10 & 3/10 & 1/10 & 1/10 & & 1/10 & \underline{8/10} & \underline{8/10} & 5/10 & 5/10 & & \underline{5/10} \\
EgoVLA           & 0/10 & 0/10 & \textbf{10/10} & 1/10 & & 0/10 & 0/10 & 0/10 & 0/10 & 0/10 & & 0/10 \\
H-RDT            & 2/10 & 0/10 & 0/10 & 0/10 & & 0/10 & 0/10 & 0/10 & 6/10 & 1/10 & & 0/10 \\
$\pi 0.5$        & \textbf{9/10} & \underline{8/10} & \underline{5/10} & \underline{5/10} & & \underline{5/10} & \underline{8/10} & \textbf{9/10} & 3/10 & 3/10 & & 3/10 \\
GR00T N1.6       & \underline{8/10} & 5/10 & 0/10 & 0/10 & & 0/10 & 7/10 & \textbf{9/10} & \textbf{8/10} & \textbf{7/10} & & 4/10 \\
\textbf{\ours (Ours)}    & \textbf{9/10} & \textbf{9/10} & \textbf{10/10} & \textbf{10/10} & & \textbf{9/10} & \textbf{9/10} & \textbf{9/10} & \underline{7/10} & \textbf{7/10} & & \textbf{6/10} \\

\midrule
\multicolumn{13}{l}{\textbf{Task 8} \hfill \textbf{Task 7}} \\
\midrule
Descriptions
& \multicolumn{6}{l|}{\makecell[l]{Pull out the tray and \\ turn to throw the chip can
into the trash}}
& \multicolumn{6}{l}{\makecell[l]{Hold the lunch bag and \\ squat down to place on
the table}} \\
\midrule
& Grasp & Pull & Walk & Drop & & \textbf{Overall}
& Hold & Turn & Squat & Put & & \textbf{Overall} \\
\midrule
Diffusion Policy & 0/10 & 0/10 & 0/10 & 0/10 & & 0/10 & 0/10 & 0/10 & 0/10 & 0/10  & & 0/10 \\
ACT              & 7/10 & 1/10 & \underline{7/10} & 0/10 & & 0/10 & 6/10 & 8/10 & 5/10 &5/10 & & 5/10 \\
InternVLA-M1     & 8/10 & \textbf{5/10} & 1/10 & 1/10 & & 1/10 & 0/10 & 0/10 & 0/10 & 0/10 & & 0/10 \\
H-RDT            & 8/10 & \underline{4/10} & 2/10 & 0/10 & & 0/10 & \underline{9/10} & \underline{9/10} & \underline{7/10} & \underline{6/10}& & \underline{6/10} \\
EgoVLA           & 0/10 & 0/10 & 0/10 & 0/10 & & 0/10 & 3/10 & 4/10 & 2/10 &2/10 & & 2/10 \\
$\pi 0.5$        & \underline{9/10} & 3/10 & 3/10 & 3/10 & & 1/10 & 3/10 & 9/10 & 2/10 & 2/10& & 2/10 \\
GR00T N1.6       & \textbf{10/10} & 1/10 & \textbf{10/10} & 3/10 & & 1/10 & 5/10 & \textbf{10/10} & 5/10 & 5/10 & & 5/10 \\
\textbf{\ours (Ours)}    & \textbf{10/10} & \textbf{5/10} & \textbf{10/10} & \textbf{9/10} & & \textbf{5/10} & \textbf{10/10} & \underline{9/10} & \textbf{9/10} & \textbf{9/10} & & \textbf{9/10} \\
\bottomrule
\end{tabular*}
\caption{\textbf{Real-World Benchmarking:} We provide a detailed report of real-world benchmarking results, including sub-task progress. Each task consists of three to five subtasks, and a trial is counted as successful only if all subtasks are completed. Boldface indicates the best performance, while underlining denotes the second-best performance.}
\label{tab:detailed_real_task}
\end{table*}

\textit{\textbf{Task 2. Spray the bowl with water, wipe clean, and fold it up:}}
The robot holds the spray bottle with its left hand and removes the cap into the bowl at the center of the desk. It then places the spray bottle back and grasps the green rag.
The robot presses the bowl with the fingers of its right hand to stabilize it, while inserting its left hand with the rag into the bowl to wipe the interior.
After cleaning, the robot places the cloth back on the table. Finally, the robot uses its right hand to stack the cleaned bowl on top of the bowl on the right.

\textit{\textbf{Task 3. Pick the bottle, turn around, and pour into cup:}}
The robot grasps the water bottle with its right hand. It then turns to the right and walks to the blue plate on the table.
The robot pours water from the bottle into the cup on the plate. Finally, the robot places the bottle on the plate.

\textit{\textbf{Task 4. Grab the can, turn and pour onto plate, push the cart forward:}}
The robot grasps the can on the table with its right hand. It then turns to the left to face the big food cart.
The robot pours the food from the can onto the plate on the cart. After pouring, the robot then places the can on the cart.
Finally, the robot grasps the handle of the cart with both hands and pushes the cart forward.

\textit{\textbf{Task 5. Push the cart, grab the grapes, and place on the plate:}}
The robot grasps the white cart containing grapes with both hands and pushes the cart toward the seated person.
The robot then grasps the grapes from the cart. It rotates its upper body to the right and places the grapes onto the plate handed by the person.

\textit{\textbf{Task 6. Put the toy into the basket, turn around, and hand it over:}}
The robot uses its left hand to place the pink dumpling toy into the small basket on the right.
It then hooks the handle of the basket with its right hand, turns around and walks toward the seated person.
Finally, the robot extends its right hand and hands the basket containing the toy to the person.

\textit{\textbf{Task 7. Hold the lunch bag and squat down to place on the table:}}
The robot holds the lunch bag on the cart with both hands.
It then rotates its upper body to the right and slowly squats down, and places the lunch bag flat on the small side table on the right.

\textit{\textbf{Task 8. Pull out the tray and turn to throw the chip can into the trash:}}
The robot grasps the chip can on the table with its right hand.
Using the index finger of its left hand, the robot inserts it into the inner tray and pulls the tray out of the can.
The robot then picks up the chip can, turns to the right, and walks toward the trash area.
Finally, the robot places the empty chip can into the recycling bin.

\subsection{Detail Evaluation Metrics}
Detailed evaluations including all the sub-task progress are provided in Table \ref{tab:detailed_real_task}.
\subsection{Deployment}
During inference, the deployment system is executed using two asynchronous threads. The policy inference thread periodically updates a shared action buffer, running at a lower frequency due to inference latency. In parallel, a low-level control thread continuously reads actions from the buffer and feeds them to the RL controller. This control thread operates at 60 Hz to ensure stable lower-body locomotion and maintain overall robot stability.

\section{More Ablation Studies}
\subsection{Effect of RTC}
In general, RTC improves action smoothness and stability, and it can reduce failures such as collisions; this might indirectly contribute to higher task success rates. Empirically, we observe that RTC slightly improves \ours \;performance.
To fully evaluate the effect of RTC, we also implement RTC on GR00T-N1.6 \cite{bjorck2025gr00t} as their code is not fully released.
The results are given in Table \ref{tab:abl-rtc-groot}.
RTC again achieves comparable performance with the baseline.

\begin{table}[h]
\centering

\begin{tabular}{lp{1cm}p{1cm}p{1cm}p{1cm}}
\toprule
GR00T-N1.6 & Pick the dumpling & Pick the hippo & Carry the box & Overall SR \\
\midrule
w/o RTC& \textbf{10/10 }& \textbf{7/10} & 9/10& \textbf{7/10} \\
w/ RTC & 6/10& \textbf{7/10} & \textbf{10/10} & 6/10 \\
\bottomrule
\end{tabular}
\caption{\textbf{GR00T with RTC.} We study the effect of RTC on the GR00T baseline.
The task consists of three steps. It achieves comparable performance on GR00T with and without RTC.}
\label{tab:abl-rtc-groot}
\end{table}

\subsection{Pre-Training on only 10\% EgoDex}
We also study the data scaling effect for pre-training. 
In this case, we use only 10\% EgoDex dataset for pre-training, we keep all protocols of post-training and fine-tuning unchanged.

\begin{table}[h]
\centering
\begin{tabular}{p{2.6cm}p{1cm}p{1cm}p{1cm}p{1cm}}
\toprule
 Experiment 1 & Pick the dumpling & Pick the hippo & Carry the box & Overall SR \\
\midrule
Baseline (\ours) & \textbf{9/10} & \textbf{9/10} & \textbf{10/10}& \textbf{8/10} \\
Variant (10\% EgoDex) & 6/10& 1/10 & 5/10 & 1/10 \\
\toprule
 Experiment 2 & Grasp bottle & Wipe the bowl & Stack up & Overall SR \\
\midrule
Baseline (\ours) & \textbf{10/10} & 9/10 & \textbf{7/10}& \textbf{7/10} \\
Variant (10\% EgoDex) & 9/10& \textbf{10/10} & \textbf{7/10} & 6/10 \\
\bottomrule
\end{tabular}
\caption{\textbf{Ablation of Pre-Training on 10\% EgoDex.} We found that using 10\% of EgoDex perform worse than the baseline \ours, demonstrating the efficacy of full EgoDex pre-training.}
\label{tab:abl-ego-10per}
\end{table}
The comparison with baseline \ours \; for two real-world experiments are given in Table \ref{tab:abl-ego-10per}.
The experiments show that using only 10\% of the EgoDex dataset leads to significantly worse performance on certain tasks and inferior overall performance.

\subsection{Pre-Training on only Humanoid Everyday}
To fully evaluate the effect of EgoDex pre-training, we pre-train only on Humanoid Everyday and keep all protocols of post-training and fine-tuning the same as baseline.
The comparisons with two baselines are given in Table \ref{tab:abl-he-}

\begin{table}[h]
\centering

\begin{tabular}{p{2.6cm}p{1cm}p{1cm}p{1cm}p{1cm}}
\toprule
 Experiment 1 & Pick the dumpling & Pick the hippo & Carry the box & Overall SR \\
\midrule
Baseline (\ours) &\textbf{9/10 }& \textbf{9/10 }& \textbf{10/10}& \textbf{8/10 }\\
Variant (HE) & \textbf{9/10}& 4/10 & \textbf{10/10} & 4/10 \\
\toprule
 Experiment 2 & Grasp bottle & Wipe the bowl & Stack up & Overall SR \\
\midrule
Baseline (\ours) & \textbf{10/10} & \textbf{9/10} & \textbf{7/10}& \textbf{7/10} \\
Variant (HE) & \textbf{10/10}& \textbf{9/10} & 4/10 & 4/10 \\
\bottomrule
\end{tabular}
\caption{\textbf{Ablation of Pre-Training on HE.} We discover that the HE variant achieves high performance on tasks that do not require fine-grained manipulation; however, it still lags behind our baseline on subtasks requiring more precise manipulation.}
\label{tab:abl-he-}
\end{table}
\subsection{Multi-Task Fine-Tuning}
\begin{figure}[h]
\centering
\includegraphics[width=0.8\columnwidth]{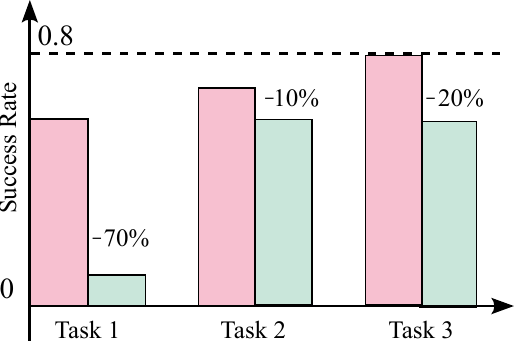}
\caption{\textbf{Multi-Task Fine-Tuning.} Joint fine-tuning (Cyan) across multiple tasks leads to a performance drop on individual tasks (Pink).}
\label{fig:abl-multi-task}
\end{figure}

We also explore the effect of multi-task fine-tuning and observed that the performance for each individual task drops compared with single task fine-tuning.
We hypothesize that multi-task training disperses the model’s learning objective and causes underfitting.
The performance comparison is reported at Fig. \ref{fig:abl-multi-task}.

\newcommand{\BW}{1.0cm} 

\newcolumntype{C}[1]{>{\centering\arraybackslash}p{#1}}
\newcommand{\W}{1.1cm}
\newlength{\TaskSpanW}
\setlength{\TaskSpanW}{\dimexpr 6\W + 12\tabcolsep \relax} 

\newcommand{\TaskHeaderL}[1]{%
\multicolumn{6}{>{\raggedright\arraybackslash}p{\TaskSpanW}|}{#1}%
}
\newcommand{\TaskHeaderR}[1]{%
\multicolumn{6}{>{\raggedright\arraybackslash}p{\TaskSpanW}}{#1}%
}




\end{document}